\def\year{2020}\relax
\documentclass[letterpaper]{article} % DO NOT CHANGE THIS
\usepackage{aaai20}  % DO NOT CHANGE THIS
\usepackage{times}  % DO NOT CHANGE THIS
\usepackage{helvet} % DO NOT CHANGE THIS
\usepackage{courier}  % DO NOT CHANGE THIS
\usepackage[hyphens]{url}  % DO NOT CHANGE THIS
\usepackage{graphicx} % DO NOT CHANGE THIS
\usepackage{graphicx}  % DO NOT CHANGE THIS
\usepackage{color}
\usepackage{booktabs}
\usepackage{amsfonts}
\usepackage{times}
\usepackage{latexsym}
\usepackage{graphicx,array}
\usepackage{url}
\usepackage{amsmath}
\usepackage{siunitx}
\usepackage{subcaption}
\usepackage[linesnumbered,ruled]{algorithm2e}
\usepackage{mathtools}
\usepackage{cleveref}[2012/02/15]% v0.18.4; 
\crefformat{footnote}{#2\footnotemark[#1]#3}
\usepackage{tabularx}
\usepackage{amsthm}

\newtheorem*{theorem*}{Theorem}
\newtheorem{proposition}{Proposition}
\newtheorem*{proposition*}{Proposition}

\newtheorem*{lemma*}{Lemma}
\DeclareMathOperator*{\KL}{KL}
\newcommand{\E}{\mathbb{E}}
\begin{document}
\title{Bivariate Beta-LSTM}
%Your title must be in mixed case, not sentence case. 
% That means all verbs (including short verbs like be, is, using,and go), 
% nouns, adverbs, adjectives should be capitalized, including both words in hyphenated terms, while
% articles, conjunctions, and prepositions are lower case unless they
% directly follow a colon or long dash
\author{Kyungwoo Song, JoonHo Jang, Seung jae Shin \and Il-Chul Moon \\ 
	Korea Advanced Institute of Science and Technology (KAIST), Korea \\
	\{gtshs2, adkto8093, tmdwo0910, icmoon\}@kaist.ac.kr}
\maketitle

\begin{abstract}
Long Short-Term Memory (LSTM) infers the long term dependency through a cell state maintained by the input and the forget gate structures, which models a gate output as a value in [0,1] through a sigmoid function. However, due to the graduality of the sigmoid function, the sigmoid gate is not flexible in representing multi-modality or skewness. Besides, the previous models lack modeling on the correlation between the gates, which would be a new method to adopt inductive bias for a relationship between previous and current input. This paper proposes a new gate structure with the bivariate Beta distribution. The proposed gate structure enables probabilistic modeling on the gates within the LSTM cell so that the modelers can customize the cell state flow with priors and distributions. Moreover, we theoretically show the higher upper bound of the gradient compared to the sigmoid function, and we empirically observed that the bivariate Beta distribution gate structure provides higher gradient values in training. We demonstrate the effectiveness of the bivariate Beta gate structure on the sentence classification, image classification, polyphonic music modeling, and image caption generation.
%, which represents the importance of each element to update the current state.
%Also, we observed that our structured, flexible gate modeling is enabled by the probability density estimation.
\end{abstract}

\section{Introduction}
%\textit{Recurrent Neural Network} (RNN) is popular for sequential data modeling because of its recurrent structure. RNNs have been widely used for text classifications\cite{DBLP:journals/corr/abs-1709-02755}, speech modeling\cite{chorowski2015attention}, sequential recommendation\cite{song2019hierarchical} and caption generations\cite{xu2015show}. %citation
One of the most commonly used Recurrent Neural Network (RNN) variants is \textit{Long Short-Term Memory} (LSTM) \cite{hochreiter1997long}, which introduces additional gate structures for controlling cell states. LSTM controls the information flow from a sequence with an input, a forget, and an output gate. The input and the forget gates decide the ratio of mixture between the current and the previous information at each time step. 
%Therefore, an effective and efficient gate mechanism would affect the performance of LSTM significantly because of dependency on the gate operations. Here, the effectiveness of the gate would be a clear and correct switching for a given pre-activation level, and its efficiency would be the learning speed of the parameters related to the pre-activation through the backpropagation.
The sigmoid function is defined to be bounded and monotonically increasing, so the sigmoid has been a popular choice for such gate mechanisms.

In spite of the prevalence of sigmoid functions, there has been a question on the utility and the efficiency of the sigmoid function used for the gates in LSTM.
%However, such boundedness and monotonicity do not guarantee the effectiveness and efficiency of the gate. 
%For instance, there are cases when we have to activate the gate with two extreme ends, 0 and 1. The traditional approach would be utilizing multiple layers to model this violation of the monotonicity, but this is also a consequence of the rigid monotonicity of the sigmoid function. 
For instance, the confined gate value range, which is narrower than the 0-1 bound, means that the majority of gate values may fall into the narrower range, and this makes that the gate values lose potential discrimination power \cite{li2018towards}. Some tried to use additional hyper-parameters to sharpen the sigmoid function, i.e., the sigmoid function with temperature parameter \cite{li2018towards}, %the temperature parameter of the Gumbel softmax \cite{45822Gumbel}, 
but these would be limited to the support for the sigmoid function without fundamental innovations. From this perspective, there are few works to probabilistically model the flexibility of the gate structure, i.e., $G^{2}$-LSTM \cite{li2018towards} with the Bernoulli distribution, but the current probabilistic model missed the graduality of the gate value change.
%, which we conjecture that it would harm the learning efficiency.
\begin{figure}
	\centering
	\begin{subfigure}[]{0.48\columnwidth}
		\includegraphics[width=\linewidth]{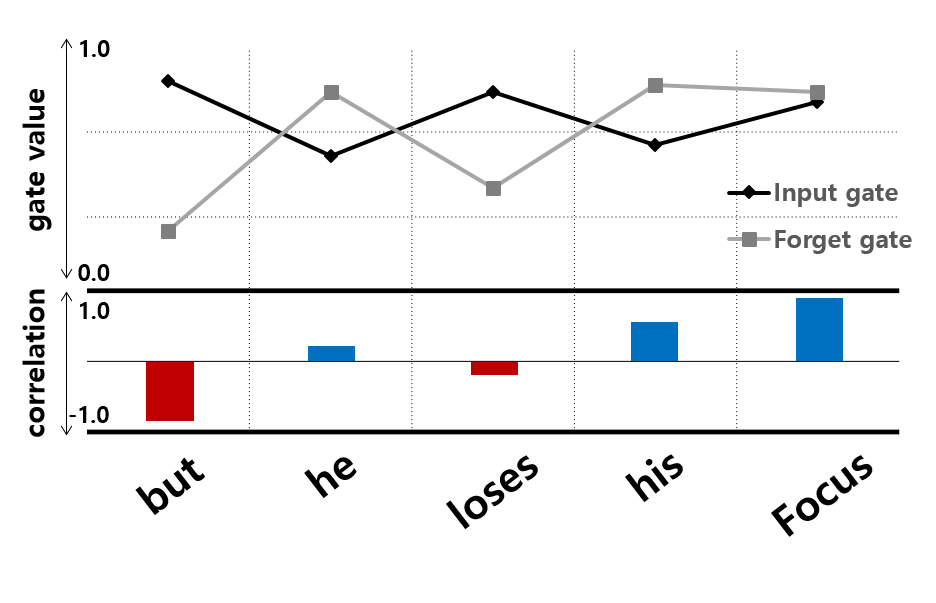}
	\end{subfigure}
	\begin{subfigure}[]{0.48\columnwidth}
		\includegraphics[width=\linewidth]{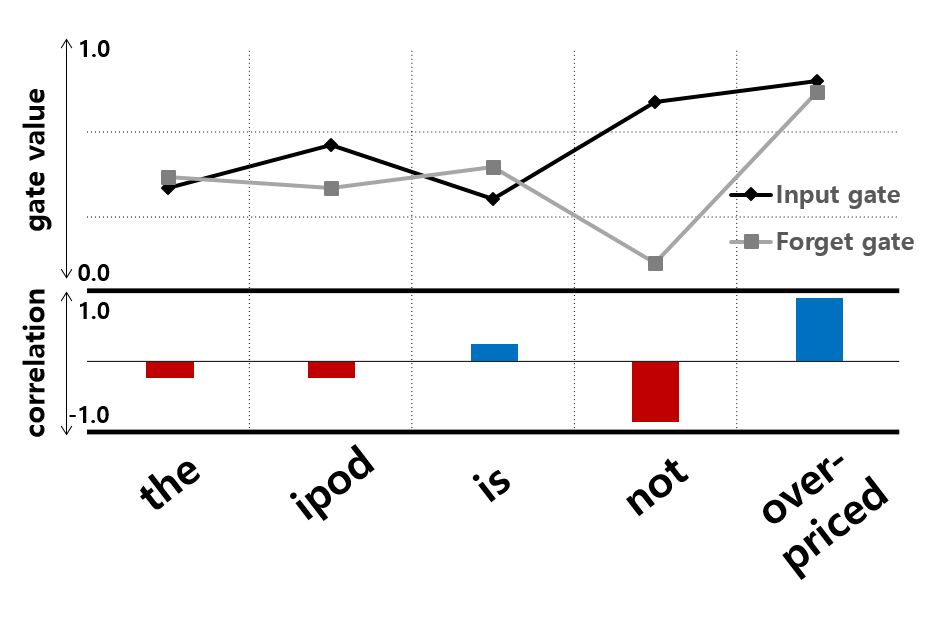}
	\end{subfigure}
	\caption{An illustrative example of the input gate, the forget gate, and their correlation for part of a given sentence in sentiment classification datasets. Blue and Red bars denote the positive and negative correlations, respectively.
		(Left) The new context starts from the word "but", so the negative correlation imposes the large input gate value and the small forget gate value to focus on the current context. 
		%Besides, the forget gate has large value on the word "his" and "focus", to convey the important word, "losses", continuously. 
		(Right) To infer the semantic meaning between the word "not" and "over-priced", the positive correlation occurs at "over-priced", so the positive correlation makes the input gate and the forget gate have large values. The high forget gate value denotes the importance of the previous context in our model.}
	\label{fig:intro_example}
\end{figure}
Moreover, it has been known that the gates could be correlated \cite{greff2017lstm}, and the performance can be improved by exploiting this covariance structure. One common conjecture is the correlation between the input and the forget gate values in LSTM. However, the structure of LSTM does not explicitly model such correlation, so its enforcing structure was handled at the technical implementation level. For instance, CIFG-LSTM \cite{greff2017lstm} enforces the negative correlation between the input and the forget gate values. CIFG-LSTM shows competitive performance with reduced parameters because of the correlation modeling. However, CIFG-LSTM enforces the strict negative correlation, -1 only; and it needs to be generalized by a model. We improve the correlation structure adaptable to datasets flexibly. % and its range lies in [-1,1] flexibly.
%mechanism should be based upon the covariance structure learning, rather than a hard specification of the correlation. 

This correlation modeling is frequently included in the data domains, such as texts and images. For text datasets, there are the syntactic and the semantic relationships between words in a sentence \cite{harabagiu2004incremental}, so the information flow to the cell structure should reflect the semantic relationship. Similarly, for image datasets, there is a relationship between pixels in a short-range, as well as pixels in a long-range within a single image \cite{kampffmeyer2019connnet}. The relation modeling is one of the effective inductive biases for deep learning models \cite{battaglia2018relational}, which can handle the property of datasets.
 
%\textcolor{red}{
In a general setting, let us assume that we prefer a large input value and a large forget gate value when both of the current and the previous information are important. Then, a positive correlation can be an effective inductive bias for modeling the idiom such as \textit{rely on}, to control the gate value, efficiently. 
For the opposite case, 
%the negative correlation can effectively control the gate value.
A negative correlation can be a good inductive bias to model the sentence "...but he loses his focus". To reflect the change of context, the input gate and the forget gate should have a large and small value, respectively, at "but" as shown in Figure \ref{fig:intro_example}. 
%As above, we can improve the LSTM by introducing the flexible gate structure with the correlation between input and forget gate.} 

We propose a bivariate Beta LSTM (bBeta-LSTM), which improves the sigmoid function in the input gate and the forget by substituting the sigmoid function with a bivariate Beta distribution. bBeta-LSTM has three advantages over the LSTM. First, the Beta distribution can represent values of [0,1] flexibly, since a Beta distribution is a generalized distribution of the uniform distribution, the power function, and the Bernoulli distribution with 0.5 probability. The Beta distribution can represent either symmetric or skewed shape by adjusting two shape parameters.
Second, the bivariate Beta distribution can represent the covariance structure of the input and the forget gates because a bivariate Beta distribution shares the Gamma random variables, which make the correlation between two sampled values. We utilized the property of the bivariate Beta distribution for modeling the input and the forget gates in bBeta-LSTM. 
%\footnote{We illustrate the gate and correlation value for the synthetic dataset in Appendix C}. 
The bivariate Beta distribution could be further elaborated by expanding the probabilistic model, i.e., adding a common prior to the input gate and the forget gate distributions. 
%Unlike other models, bBLSTM based one five Gamma distribution can represent the correlation between -1 and 1.
Third, the bivariate Beta distribution can alleviate the gradient vanishing problem of LSTM. Under a certain condition, we verify that the derivative of gates in bBeta-LSTM is greater than those of LSTM, experimentally and theoretically. 

\begin{figure}[t!] % modify to 2 by 2
	\centering
	\begin{subfigure}{.48\columnwidth}
		\centering
		\includegraphics[width=\linewidth]{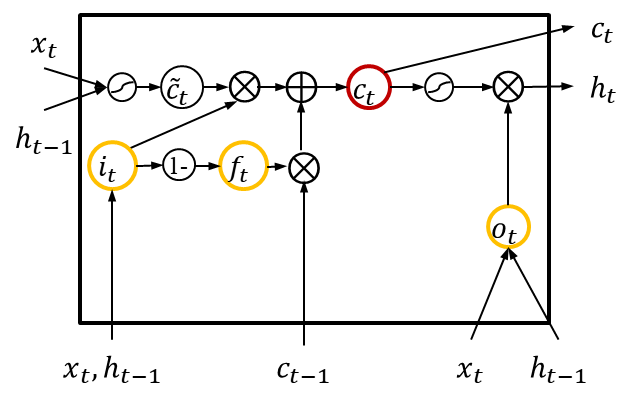}
		\caption{CIFG-LSTM}
		\label{fig:CIFGLSTM_cell}
	\end{subfigure}\hfil
	\begin{subfigure}{.48\columnwidth}
		\centering
		\includegraphics[width=\linewidth]{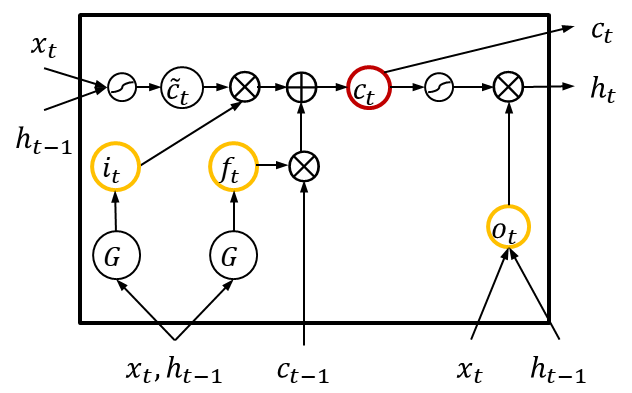}
		\caption{$G^{2}$-LSTM}
		\label{fig:G2LSTM_cell}
	\end{subfigure}
	\caption{The cell structure of CIFG-LSTM and $G^{2}$-LSTM}
\end{figure}

%\begin{figure*}[]
%	\centering
%	\begin{minipage}[b]{0.24\linewidth}
%		\centering
%		\includegraphics[width=\linewidth]{CIFGLSTM_cell.png}
%		\caption{CIFG-LSTM}\label{fig:CIFGLSTM_cell}
%	\end{minipage}
%	\begin{minipage}[b]{0.24\linewidth}
%		\centering
%		\includegraphics[width=\linewidth]{G2LSTM_cell.png}
%		\caption{$G^{2}$-LSTM}\label{fig:G2LSTM_cell}
%	\end{minipage}
%	\begin{minipage}[b]{0.48\linewidth}
%		\begin{align}
%			&i_{t} = G(W_{xi}x_{t}+W_{hi}h_{t-1}+b_{i},\tau) \\
%			&f_{t} = G(W_{xf}x_{t}+W_{hf}h_{t-1}+b_{f},\tau) \\
%			&\widetilde{c}_{t} = tanh(W_{xc}x_{t}+W_{hc}h_{t-1}+b_{c}) \\
%			&c_{t} = f_{t} \odot c_{t-1} + i_{t} \odot \widetilde{c}_{t} \label{eq:G2_ct}\\
%			&o_{t} = \sigma(W_{xo}x_{t}+W_{ho}h_{t-1}+b_{o}) \\
%			&h_{t} = o_{t} \odot tanh(c_{t})
%		\end{align}
%	\end{minipage}
%\end{figure*}\hfill

%\section{Preliminary: Stochastic Gate Mechanism in Recurrent Neural Networks}
\section{Preliminary: Stochastic Gate in RNN}
%\subsection{}
%% stochastic
%RNN has been successfully utilized for time series modeling because of its modeling on dynamic structures. 
Since RNN is a deterministic model, it is difficult to prevent overfitting, and it is infeasible to generate diverse outputs. Therefore, multiple methods were explored to model the stochasticity in sequence learning. First, dropout methods for RNN \cite{gal2016theoretically,DBLP:journals/corr/abs-1904-09816} demonstrated that stochastic masking could improve its generalization. Second, latent variables were a good combination with the RNN structure, such as Variational RNN (VRNN) \cite{chung2015recurrent} and Variable Hierarchical Recurrent Encoder Decoder (VHRED) \cite{serban2017hierarchical}. Third, the gate mechanisms, which are extensively used in RNN variants due to the vanishing gradient, can be substituted with probabilistic models.

%% gate structure
When we investigate further on the gate mechanism, there have been efforts in reducing the number of gates \cite{sru}, enabling a gate structure to be a complex number \cite{NIPS2018_8253}, correlating gate structures \cite{greff2017lstm}. 
%The probabilistic gate mechanism has been recently introduced, and there are a few kinds of research. 
% Not a good paper. If we have to reduce the size, this is the first part to be removed.
%For example, Gaussian gated LSTM\cite{thornton2018gaussian} introduces an additional gate at the outside of a cell structure to reduce the number of state updates. The additional gate, Gaussian gate, controls the updates of states, and the gate plays a similar role like skip modeling. Having said that, this gate was not truly probabilistic given that it lacks statistical inference to model the parameters of the Gaussian distribution, so eventually, the gate only becomes a kernel function for static parameters. 
For instance, Gumbel Gate LSTM ($G^{2}$-LSTM) \cite{li2018towards} replaces the sigmoid function of the input and the forget gates with Bernoulli distributions. The Bernoulli gates in $G^{2}$-LSTM turns the continuous gate values to be the binary value of 0 or 1. 
%This work is interesting because of its reparameterization effort to model the Bernoulli distribution. 
This work can be expanded to incorporate a continuous spectrum, a multi-modality, and stochasticity, at the same time. Furthermore, $G^{2}$-LSTM uses a Gumbel-Softmax, which remains in the realm of sigmoid gates, so the limitations discussed in the Introduction are still applicable. Figure \ref{fig:G2LSTM_cell} and below equations enumerate the information flow in $G^{2}$-LSTM, and $G$ is the Gumbel-softmax function with a temperature parameter of $\tau$.
\begin{align}
	&i_{t} = G(W_{xi}x_{t}+W_{hi}h_{t-1}+b_{i},\tau) \\
	&f_{t} = G(W_{xf}x_{t}+W_{hf}h_{t-1}+b_{f},\tau) \\
	&\widetilde{c}_{t} = tanh(W_{xc}x_{t}+W_{hc}h_{t-1}+b_{c}) \\
	&c_{t} = f_{t} \odot c_{t-1} + i_{t} \odot \widetilde{c}_{t} \label{eq:G2_ct}\\
	&o_{t} = \sigma(W_{xo}x_{t}+W_{ho}h_{t-1}+b_{o}) \\
	&h_{t} = o_{t} \odot tanh(c_{t})
\end{align}
When we consider a stochastic expansion on gate mechanisms, it is natural to structure the random variables with conditional independence and priors. For example, the input and the forget gates are both related to the cell state in the LSTM cell so that we may conjecture their correlations through a common cause prior. To our knowledge, CIFG-LSTM in Figure \ref{fig:CIFGLSTM_cell} is the first model to introduce a structured input and forget gate modeling by assuming $f_{t}=1-i_{t}$. This hard assignment is not a flexible correlation modeling, so this can be further extended by adopting the flexible probabilistic gate mechanism. Our source code is available at https://github.com/gtshs2/BetaLSTM.

\section{Methodology}
\begin{figure*}[t!] % modify to 2 by 2
	\centering
	\begin{subfigure}{.48\columnwidth}
		\centering
		\includegraphics[width=\linewidth]{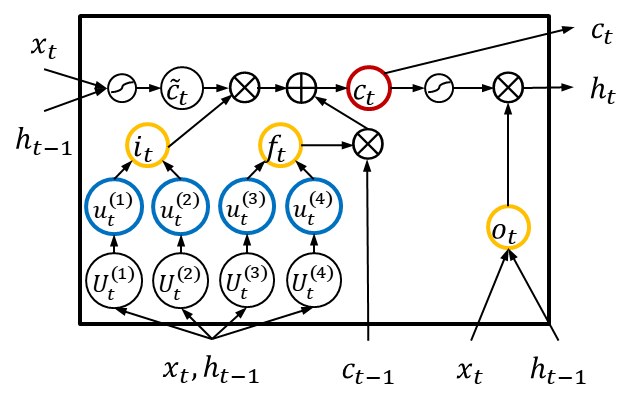}
		\caption{Beta-LSTM} 
	\end{subfigure}\hfil
	\begin{subfigure}{.48\columnwidth}
		\centering
		\includegraphics[width=\linewidth]{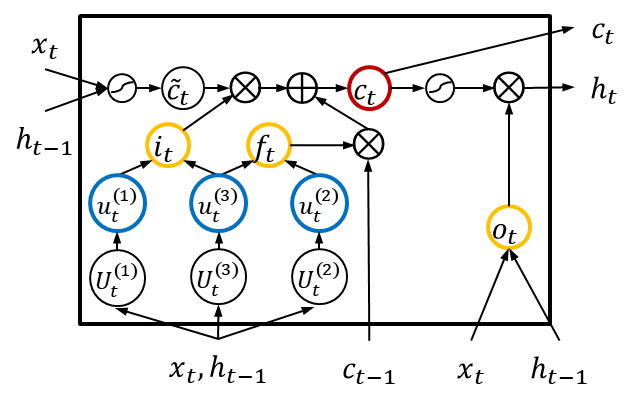}
		\caption{bBeta-LSTM(3G)} 
	\end{subfigure}
	\begin{subfigure}{.48\columnwidth}
		\centering
		\includegraphics[width=\linewidth]{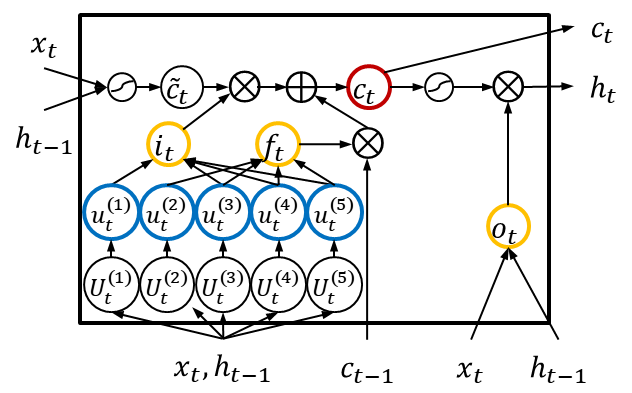}
		\caption{bBeta-LSTM(5G)} 
	\end{subfigure}\hfil
	\begin{subfigure}{.48\columnwidth}
		\centering
		\includegraphics[width=\linewidth]{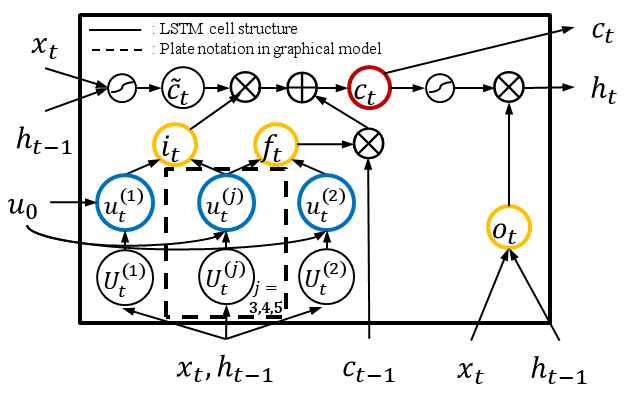}
		\caption{bBeta-LSTM(5G+p)}
		\label{fig:bBLSTM(5G)_prior_cell}
	\end{subfigure}
	\caption{The cell structure of our proposed models. The red and yellow circle denotes a cell state and gates, respectively. The blue circle represents random variables that follow the Gamma distribution. The input and the forget gates in bBeta-LSTM(5G) and bBeta-LSTM(5G+p) shares random variables, $u^{(3)},u^{(4)},u^{(5)}$ and prior $u^{(0)}$.}
	\label{fig:our_cell_structure}
\end{figure*}

% https://pytorch.org/docs/stable/_modules/torch/distributions/gamma.html
% https://en.wikipedia.org/wiki/Gamma_distribution
% u ~ Gamma(shape,rate) = Gamma(concentration,1/scale)
First, we improve the LSTM to have more flexible gate values, which can represent skewness and multi-modality by modeling the input and the forget gate as a Beta distribution. Second, we extend the Beta distribution to incorporate the correlation between the input gate and the forget gate with the bivariate Beta distribution. Third, we introduce the prior distribution to the gate structure to keep the stochasticity and handle the mutual dependency.
Our probabilistic gate model resides in a neural network cell, as Figure \ref{fig:our_cell_structure}.

\subsection{Beta-LSTM}
We propose a Beta-LSTM that embeds independent Beta distributions on the input and the forget gates,  instead of the sigmoid function. We construct each Beta distribution with two Gamma distributions to apply the reparametrization technique.
\begin{align}
	&U_{t}^{(j)} = g_{j}(x_{t},h_{t-1}), j=1,...,4 \label{eq:U}\\
	&u_{t}^{(j)} \sim \text{Gamma}(U_{t}^{(j)},1), j=1,..,4 \label{eq:u}\\
	&i_{t} = \frac{u_{t}^{(1)}}{u_{t}^{(1)}+u_{t}^{(2)}}, f_{t} = \frac{u_{t}^{(3)}}{u_{t}^{(3)}+u_{t}^{(4)}}
\end{align}
We formulate $U_{t}^{(j)}$, the shape parameter of a Gamma distribution; as a function, $g_j$ of the current input $x_{t}$ and the previous hidden state $h_{t-1}$. We omit the amortized inference on the rate parameter of a Gamma distribution by setting it as a constant of 1. Each $g_{j}$ can be a multi-layered perceptron (MLP) that combines $x_{t}$ and $h_{t-1}$. 
%Given the shape parameter's constraint of being positive, the final output of the MLP can be gained by either $softplus$ or $Relu$.

As we follow the reparameterization technique of optimal mass transport (OMT) gradient estimator \cite{jankowiak2018pathwise} which utilize the implicit differentiation, we can compute the stochastic gradient of random variable $u_{t}^{(j)}$ with respect to $U_{t}^{(j)}$ efficiently, without inverse CDF.
%read Appendix A for details.

%linear function, $W_{x_{u_i}}x_{t} + W_{h_{u_{i}}} h_{t} + b_{u_{i}}$, followed by activation function such as $softplus$ or $relu$. 
%The role of activation function for each $g_{i}$ is to restrict the $U_i$ as a positive value, and it is used as the Gamma distribution parameters.
%Beta-LSTM improve the input and forget gates in LSTM to have more flexible and stochastic value.

\subsection{\textit{bivariate} Beta-LSTM}
Beta-LSTM improves LSTM to have more flexible input and forget gate values, but these inputs and forget gates are modeled independently, which is the same as LSTM. However, as we surveyed in the above, there is a growing interest in modeling the correlation of the gate values. To consider the correlation efficiently, we further extended Beta-LSTM to have a structured gate modeling. We adopt the bivariate Beta distribution to reflect the correlation between input and forget gates by maintaining the flexibility of the Beta distribution. We can construct a bivariate Beta distribution with three independent random variables which follow a Gamma distribution, independently \cite{OLKIN2003407}. We name the bivariate Beta LSTM with three Gamma distributions as bBeta-LSTM(3G). The formulation of $U_{t}^{(j)}$ and $u_{t}^{(j)}$ is same within Equation \ref{eq:U},\ref{eq:u} for all $j$.
\begin{align}
	%&U_{t}^{(j)} = g_{j}(x_{t},h_{t-1}), j=1,...,3 \\
	%&u_{t}^{(j)} \sim Gamma(U_{t}^{(j)},1), j=1,..,3 \\
	&i_{t} = \frac{u_{t}^{(1)}}{u_{t}^{(1)}+u_{t}^{(3)}}, f_{t} = \frac{u_{t}^{(2)}}{u_{t}^{(2)}+u_{t}^{(3)}} \label{eq:3G_input_forget}
\end{align}
The bivariate Beta distribution utilizes Gamma random variables to handle the correlation between the input and the forget gate values, but bBeta-LSTM(3G) can only model the positive correlation between 0 and 1 \cite{OLKIN2003407}. 
In practice, for example, natural language processing, the input, and the forget gates might show either a positive or negative correlation in cases. Sequential correlated words, i.e., idioms or phrases, would prefer a positive correlation because the cell state should include both previous and current information. 
On the contrary, if a new important context starts, unlike the previous context, the cell state should disregard the previous information and adapt the current information. The latter case will require a negative correlation, but bBeta-LSTM(3G) lacks this functionality.
%The positive correlation between input and forget gates might be helpful to handle the phrase or idiom. However, the number of hidden neuron in LSTM is fixed, and it is hard to remember all of the previous information. Especially, we need to remove the previous hidden when the current inputs are critical information independent with the previous information. We can improve bBeta-LSTM by imposing a negative correlation modeling between input and forget as well as positive correlation modeling.  

We extend the bivariate Beta distribution in bBeta-LSTM(3G) to be bBeta-LSTM(5G) that uses a bivariate Beta distribution with a more flexible covariance structure. bBeta-LSTM(5G) consists of five random variables following a Gamma distribution, and bivariate Beta distribution with five random variables can handle both negative and positive correlation \cite{ARNOLD20111194}. bBeta-LSTM(5G) is a generalized model of CIFG-LSTM with a probabilistic covariance model. The formulation of $U_{t}^{(j)}$ and $u_{t}^{(j)}$ is same within Equation \ref{eq:U},\ref{eq:u} for all $j$.
%Our proposed model, bivariate Beta LSTM with five Gamma (bBeta-LSTM(3G)) can handle the both of negative and positive correlation between -1 and 1. 
%bBeta-LSTM(5G) is a generalization of CIFG-LSTM in terms of correlation modeling.
%We propose bBeta-LSTM(5G) which utilize the bivariate Beta distribution with five Gamma distribution.
\begin{align}
	%&U_{t}^{(j)} = g_{j}(x_{t},h_{t-1}), j=1,...,5 \\
	%&u_{t}^{(j)} \sim Gamma(U_{t}^{(j)},1), j=1,..,5 \\
	&i_{t} = \frac{u_{t}^{(1)}+u_{t}^{(3)}}{u_{t}^{(1)}+u_{t}^{(3)}+u_{t}^{(4)}+u_{t}^{(5)}} \label{eq:5G_input} \\ 
	&f_{t} = \frac{u_{t}^{(2)}+u_{t}^{(4)}}{u_{t}^{(2)}+u_{t}^{(3)}+u_{t}^{(4)}+u_{t}^{(5)}} \label{eq:5G_forget}
\end{align}

Another advantage of using a bivariate Beta distribution as an activation function is resolving the gradient vanishing problem of LSTM. We provide a proposition that the gradient value of a gate value in bBeta-LSTM(5G) with respect to the gate parameter is larger than that of LSTM under a certain condition.

\begin{proposition}
%	(Given the proof in Appendix D.)
Let $i_{t}^{a}(V_{t})$ and $i_{t}^{b}(U_{t}^{(1:5)})$ be the input gate of LSTM and bBeta-LSTM(5G) respectively, where $V_{t}$ and $U_{t}^{(1:5)}$ are input of each gate. Suppose that $u_{t}^{(j)}<0.8$ or $8<U_{t}^{(j)}$ which satisfy the $|u_{t}^{(j)}-U_{t}^{(j)}|\leq \delta \cdot U_{t}^{(j)}$ for all $j$ and $\delta>0$. Then, for the fixed $u_{t}^{(3:5)}$, $\frac{\partial i_{t}^{b}(U_{t}^{(1:5)})}{\partial U_{t}^{(1)}}|_{U_{t}^{(1)}=0.5}$ has greater maximum value than $\frac{\partial i_{t}^{a}(V_{t})}{\partial V_{t}}$.
\end{proposition}

bBeta-LSTM(5G) considers the input gate and the forget gates as a bivariate Beta distribution, so bBeta-LSTM(5G) represents the flexible gate structure with either positive or negative correlation. Besides, stochasticity in bivariate Beta distribution can alleviate the overfitting. However, the bivariate Beta random variables with five Gamma (Eq.\ref{eq:5G_input},\ref{eq:5G_forget}), can has a lower variance than the bivariate Beta random variables with three Gamma (Eq.\ref{eq:3G_input_forget}), and the variance can be near to zero under the no regularization.  The low variance can limit the advantage of stochasticity \cite{dieng2018noisin}, and we need additional prior model to regularize the gamma random variable $u_{t}^{(i)}$.

\begin{proposition}
%(Given the proof in Appendix D.)
	If all of $u_{t}^{(j)}$ have same fixed value for $j=1,...,5$, the variance of $i_{t}$ ($f_{t}$) in bBeta-LSTM(5G) is less than the variance of $i_{t}$ ($f_{t}$) in bBeta-LSTM(3G).
\end{proposition}

\subsection{\textit{bivariate} Beta-LSTM with Structured Prior Model}
Hierarchical Bayesian modeling can impose uncertainty on a model as well as a mutual dependence between variables. %The uncertainty and mutual dependency improve the model to behave robustly and generalize well. 
bBeta-LSTM(5G) has a component of probabilistic modeling, and it is easy to incorporate a prior distribution to the likelihood of the gate value.  We propose bBeta-LSTM(5G) with prior, denoted by bBeta-LSTM(5G+p), and we optimize bBeta-LSTM(5G+p) by maximizing the log marginal likelihood of the target sequence $y_{1:T}$ in Equation  \ref{eq:log_marginal_likelihood}, see Figure \ref{fig:bBLSTM(5G)_prior_cell} which combines the neural network gates and the random variables.
%\footnote{We release our code on https://github.com/AnoPaperSub/bBLSTM}.
Given the latent dimension of the prior, we utilize the variational method, and we optimize the evidence lower bound (ELBO) \cite{VAE2014} in Equation \ref{eq:ELBO} with a variational distribution, $q$, which is a feed-forward neural network with the current input $x_{t}$ and the previous hidden $h_{t-1}$.

%However, the direct maximization of the log marginal likelihood is intractable, so we approximate the log marginal likelihood with the evidence lower bound (ELBO) \cite{ELBO1999,VAE2014} in Equation \ref{eq:ELBO} with a variational distribution, $q$, which is a feed-forward neural network with the current input $x_{t}$ and the previous hidden $h_{t-1}$.
\begin{align}
\log{p}(y_{1:T}) &= \log \int \prod_{t=1}p(u_{t}^{(1:5)})p(y_{t}|u_{t}^{(1:5)})du_{t}^{(1:5)}
\label{eq:log_marginal_likelihood}\\
\mathcal{L}_{ELBO} &= \sum_{t=1}^{T}\E_{q(u_{t}^{(1:5)}|x_{t},h_{t-1})}[p(y_{t}|u_{t}^{(1:5)})] \nonumber \\
&- \KL[q(u_{t}^{(1:5)}|x_{t},h_{t-1}) \, \Vert \, p(u_{t}^{(1:5)})] \label{eq:ELBO}
\end{align}
%The prior distribution in Equation \ref{eq:log_marginal_likelihood},\ref{eq:ELBO} becomes a distribution with either constant parameters or the parameters inferred by a neural network. VRNN \cite{chung2015recurrent} and VHRED \cite{serban2017hierarchical}, which are the variational recurrent model for the context modeling, set up their priors as neural networks depending on the previous input. Similarly, 
We model a prior distribution of $p(u_{t}^{(1:5)})$, as a Gamma distribution which is a conjugate distribution of the Gamma distribution of $u_{t}^{(1:5)}$ in Equation \ref{eq:ELBO}. A Gamma distribution takes two parameters, which represent shape and rate, and our framework enables learning of the two parameters with an inference network.

\subsubsection{Domain Customized Structured Prior}
The prior on the gate in bBeta-LSTM(5G+p) is extended to incorporate other probabilistic generative models, such as Latent Dirichlet Allocation (LDA) \cite{blei2003latent} or word vector, i.e., Glove or Word2Vec. Considering the input and the forget gates reside in a LSTM cell at a certain time, $t$, the prior can be better informed by a global context extracted from $x_{1:T}$. To demonstrate this capability, we adapted Equation \ref{eq:ELBO} to be as below.
\begin{align}
&\sum_{t=1}^{T}\E_{q(u_{t}^{(1:5)}|x_{t},h_{t-1})}[p(y_{t}|u_{t}^{(1:5)})] \nonumber \\
&- \lambda\{\KL[q(u_{t}^{(3:5)}|x_{t},h_{t-1}) \, \Vert \, p(u_{t}^{(3:5)})] \nonumber \\
&+\KL[q(u_{t}^{(1:2)}|x_{t},h_{t-1}) \, \Vert \, p(u_{t}^{(1:2)}|\beta_{t-1},\beta_{t})]\}.
\end{align}
Here, $\lambda$ is the weight of the prior regularization, to balance the likelihood and the $\KL$ regularization term, and $\beta_{t}$ is the topic probability of a word at time $t$ in the sequence, which follows the definition in the original LDA. If we impose that $\beta_{t}$ is related prior to $u_{t}^{(1)}$ and $u_{t}^{(2)}$, which contribute to $i_t$ and $f_t$, respectively; we can directly reflect the global context to compute the input and the forget gates.
To reflect the global context adequately, we compare the similarity between the topic proportion of the previous word and the current word with the radial basis kernel function (RBF).
%, $k_{RBF}$,  $p(u_{t}^{(1:2)}|\beta_{t-1},\beta_{t})=\text{Gamma}(k_{RBF}(\beta_{t},\beta_{t-1}),1)$.
Under the prior modeling, we can also compute the input gate and the forget gates by reflecting the similarity of global topic proportions between sequential words. If the sequential words share a similar topic proportion, similar semantics, the prior with RBF kernel encourages $u_{t}^{(1:2)}$ to have a large value. This makes the large input and the large forget gate values, so the gating mechanisms handle the semantic composition of the previous input and the current input. 
While we used a pre-trained LDA model, we can learn the parameters of our proposed models and LDA parameter simultaneously. Additionally, $\beta_{t}$ can be substituted by a word vector, i.e., Glove.
\begin{figure}
	\centering
	\begin{subfigure}[]{0.48\columnwidth}
		\includegraphics[width=\linewidth]{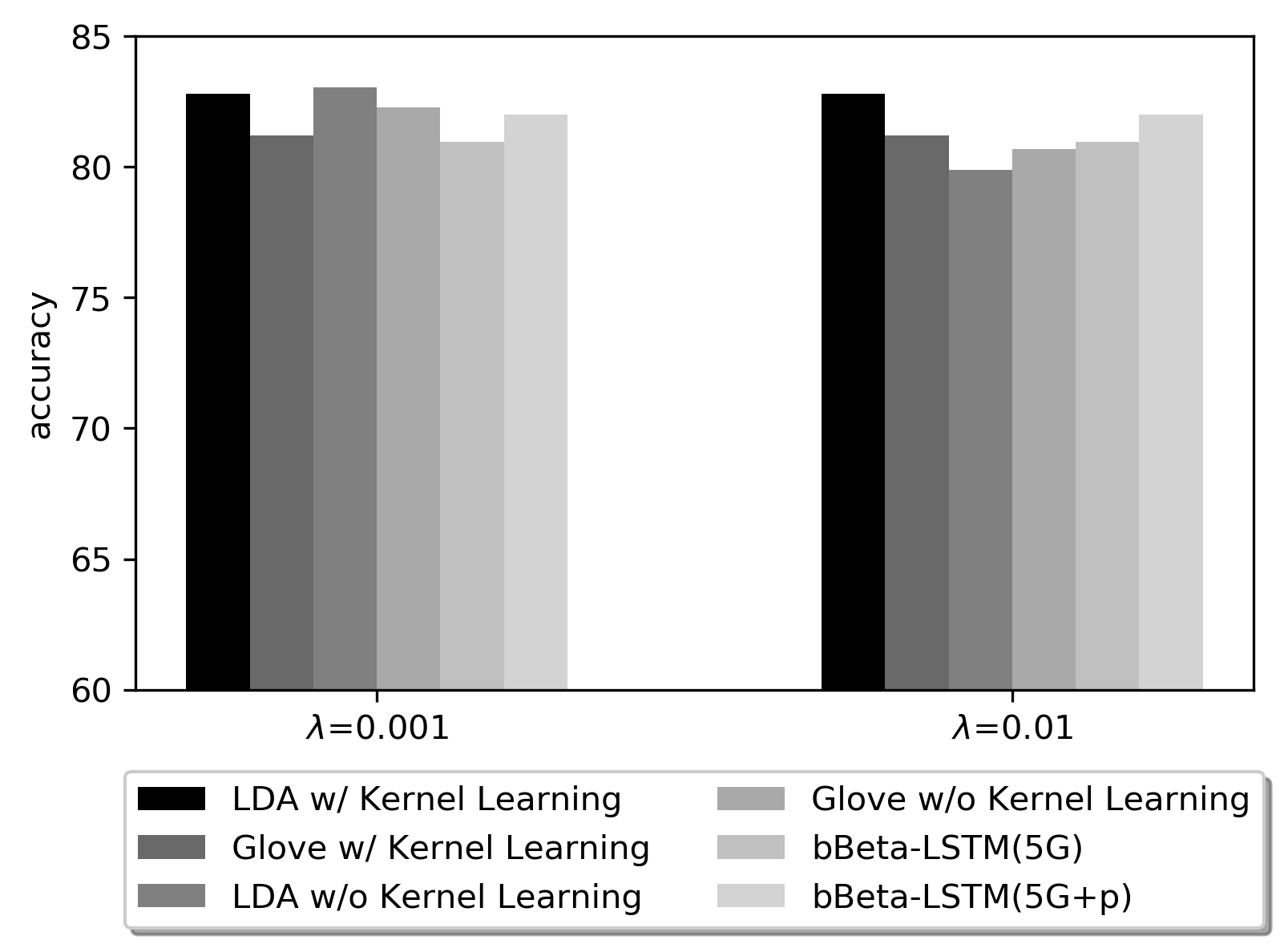}
	\end{subfigure}
	\begin{subfigure}[]{0.48\columnwidth}
		\includegraphics[width=\linewidth]{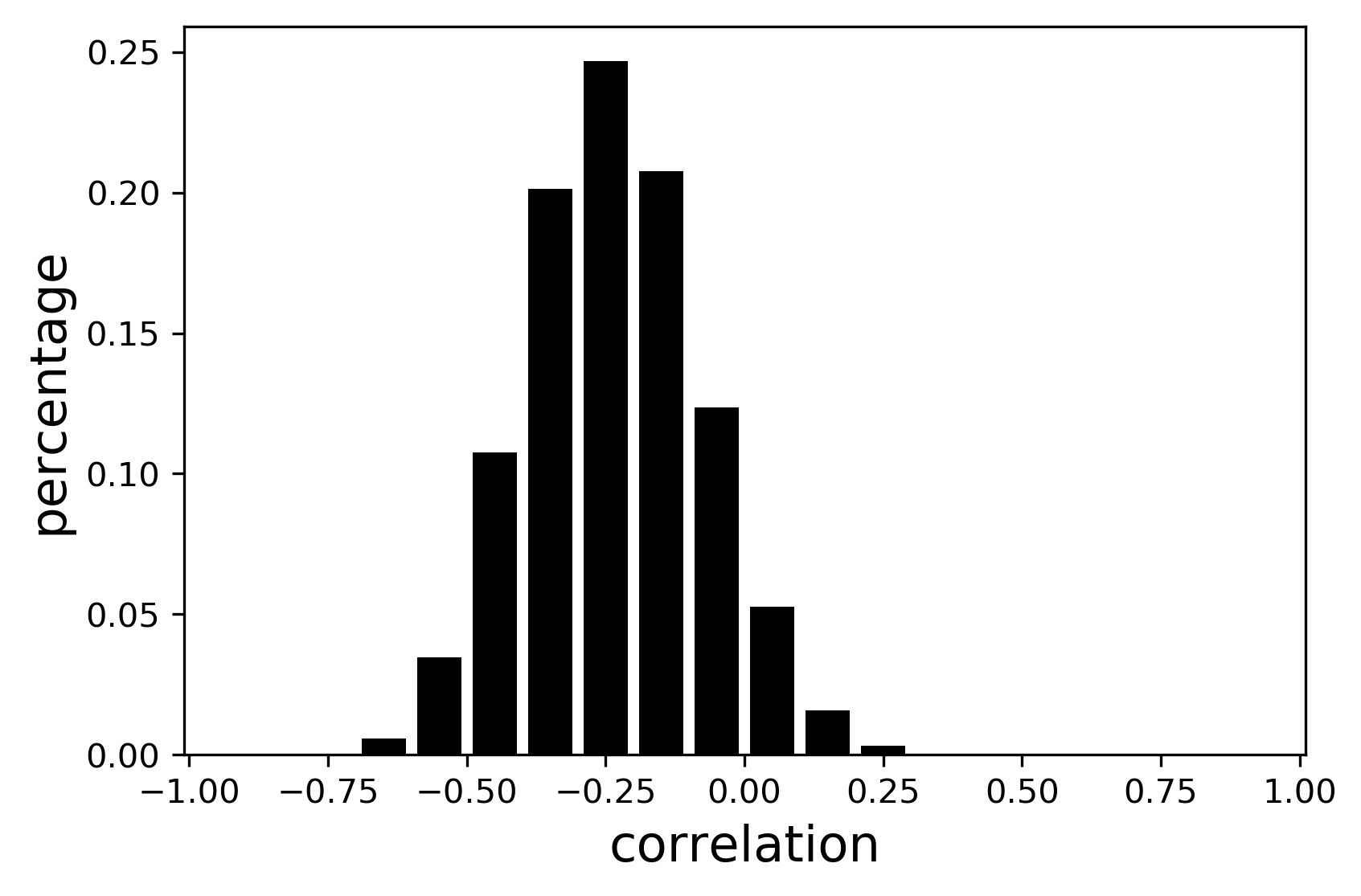}	
	\end{subfigure}
	\caption{(Left) The sentiment classification accuracy of bBeta-LSTM variants with or without prior learning on a $0_{th}$ fold of CR dataset. (Right) The correlation of bBeta-LSTM(5G+p) on CR dataset.}
	\label{fig:prior_learning_corr_cr_mr}
\end{figure}
%Finally, we assume $p(u_{t}^{(1:2)}|\beta_{t-1},\beta_{t})=\text{Gamma}(k_{RBF}(\beta_{t},\beta_{t-1}),1)$. Since a radial basis kernel function, $k_{RBF}$, is used, its length and scale parameters is also learnable. The model on $p(u_{t})$ can be adapted to domains, and our modeling motivation is capturing the significant word semantic compositions to influence the input and the forget gate outputs. It should be noted that $k_{RBF}$ can be replaced by MLP. Additionally, $\beta$ can be substituted by a word vector, i.e. Glove. 

Figure \ref{fig:prior_learning_corr_cr_mr} (Left) illustrates three insights. First, the strength of the prior should be limited by $\lambda$. Second, the prior with LDA is generally better than the prior with a static parameter, bBeta-LSTM(5G+p). Third, it is important to learn the inference model, i.e. the kernel hyperparameters used for the parameter of $p(u_{t}^{(1:2)}|\beta_{t-1},\beta_{t})$. 

\begin{table*}[h!]
	%	\fontsize{9.5}{12}\selectfont
	\centering
	\begin{tabular}{lcccccc}
		\toprule
		\bfseries Models & \bfseries CR & \bfseries SUBJ & \bfseries MR & \bfseries TREC & \bfseries MPQA & \bfseries SST\\
		\midrule
		LSTM & 82.91$\pm$2.40 & 92.58$\pm$0.84 & 80.37$\pm$0.98& 94.42$\pm$1.07 & 89.38$\pm$0.55 &88.13$\pm$0.67\\
		CIFG-LSTM & 83.28$\pm$1.79 & 92.65$\pm$0.86& 79.86$\pm$0.91& 94.00$\pm$0.78& 89.14$\pm$0.91 & 87.63$\pm$0.46\\
		$G^{2}$-LSTM  & 83.31$\pm$1.66 & 92.69$\pm$0.78& 80.13$\pm$1.10& 94.68$\pm$0.37& 89.34$\pm$0.54& 88.36$\pm$0.96\\
		\midrule
		Beta-LSTM  & 84.45$\pm$1.87 & 93.25$\pm$0.88& 81.12$\pm$0.93& 94.38$\pm$0.64& 89.66$\pm$0.49& 88.68$\pm$0.67\\
		bBeta-LSTM(3G)  & 83.63$\pm$2.14 & 93.23$\pm$0.78& 81.47$\pm$0.92& 94.28$\pm$0.48 & 89.41$\pm$0.91& 88.23$\pm$0.67\\
		bBeta-LSTM(5G) & 84.48$\pm$1.96 & 92.87$\pm$0.74& 81.05$\pm$1.05& 94.30$\pm$0.69& 89.42$\pm$0.66& 88.89$\pm$0.46\\
		bBeta-LSTM(5G+p) & \textbf{84.66}$\pm$2.42 & \textbf{93.25}$\pm$0.85&\textbf{81.59}$\pm$0.91&\textbf{94.80}$\pm$0.47&\textbf{89.66}$\pm$0.44&\textbf{88.94}$\pm$0.43 \\
		\midrule
		SRU(8-layers) & 87.00$\pm$2.24 & 93.76$\pm$0.61 & 83.14$\pm$1.53 & 94.52$\pm$0.34 & 90.39$\pm$0.55 & 89.58$\pm$0.46\\
		\hspace{1pt} + bBeta-LSTM(5G+p)  & \textbf{87.21}$\pm$0.78 & \textbf{93.97}$\pm$0.55 & \textbf{83.51}$\pm$1.42 & \textbf{94.80}$\pm$0.47 & \textbf{90.44}$\pm$0.79 & \textbf{89.72}$\pm$0.31\\
		\bottomrule
	\end{tabular}
	\caption{Test accuracies on sentence classification task.}
	\label{table:sentence}
\end{table*}
\begin{figure*}[h]
	\centering
	\begin{subfigure}{.48\columnwidth}
		\centering
		\includegraphics[width=\linewidth]{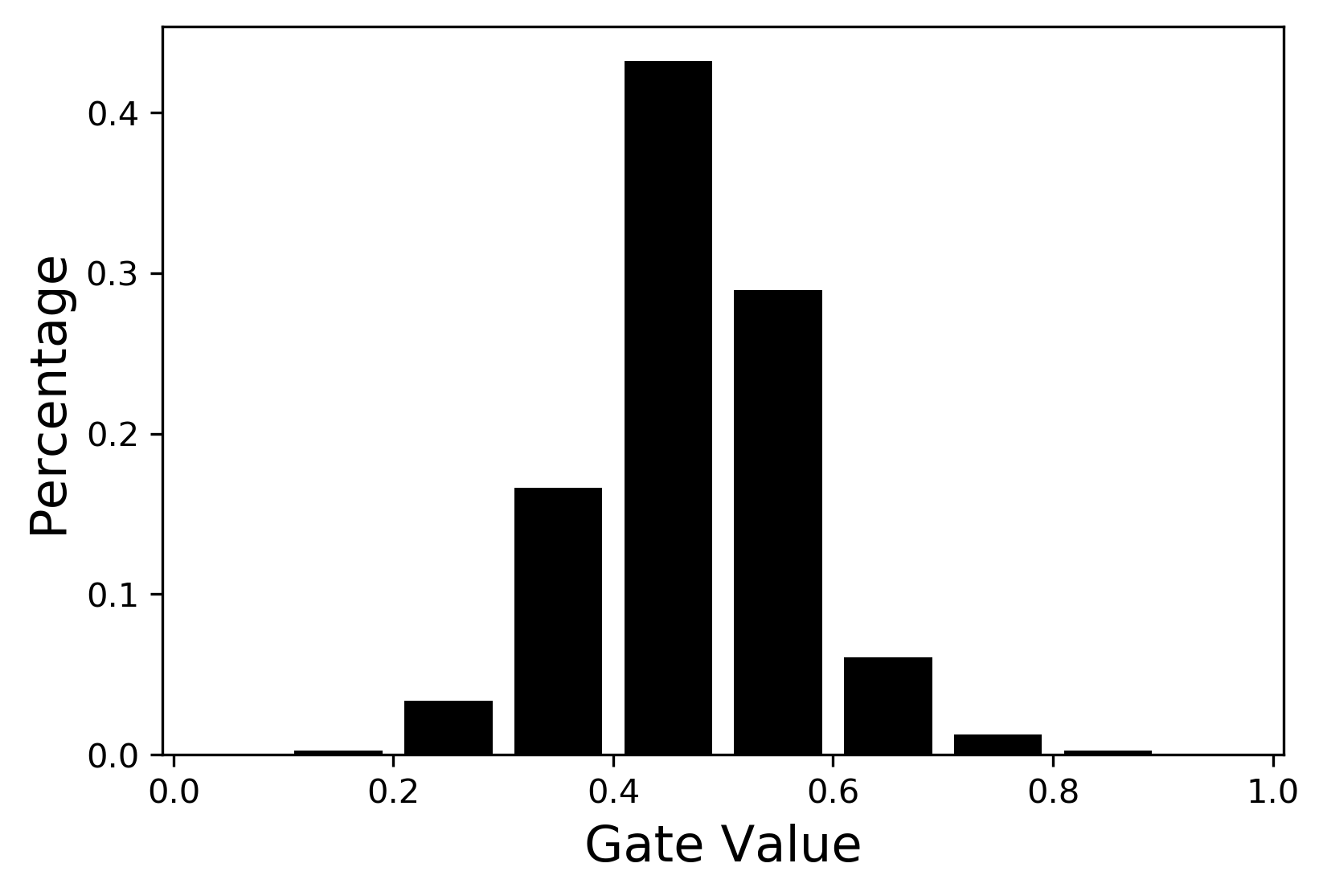}
		\caption{LSTM}
	\end{subfigure}\hfil
	\begin{subfigure}{.48\columnwidth}
		\centering
		\includegraphics[width=\linewidth]{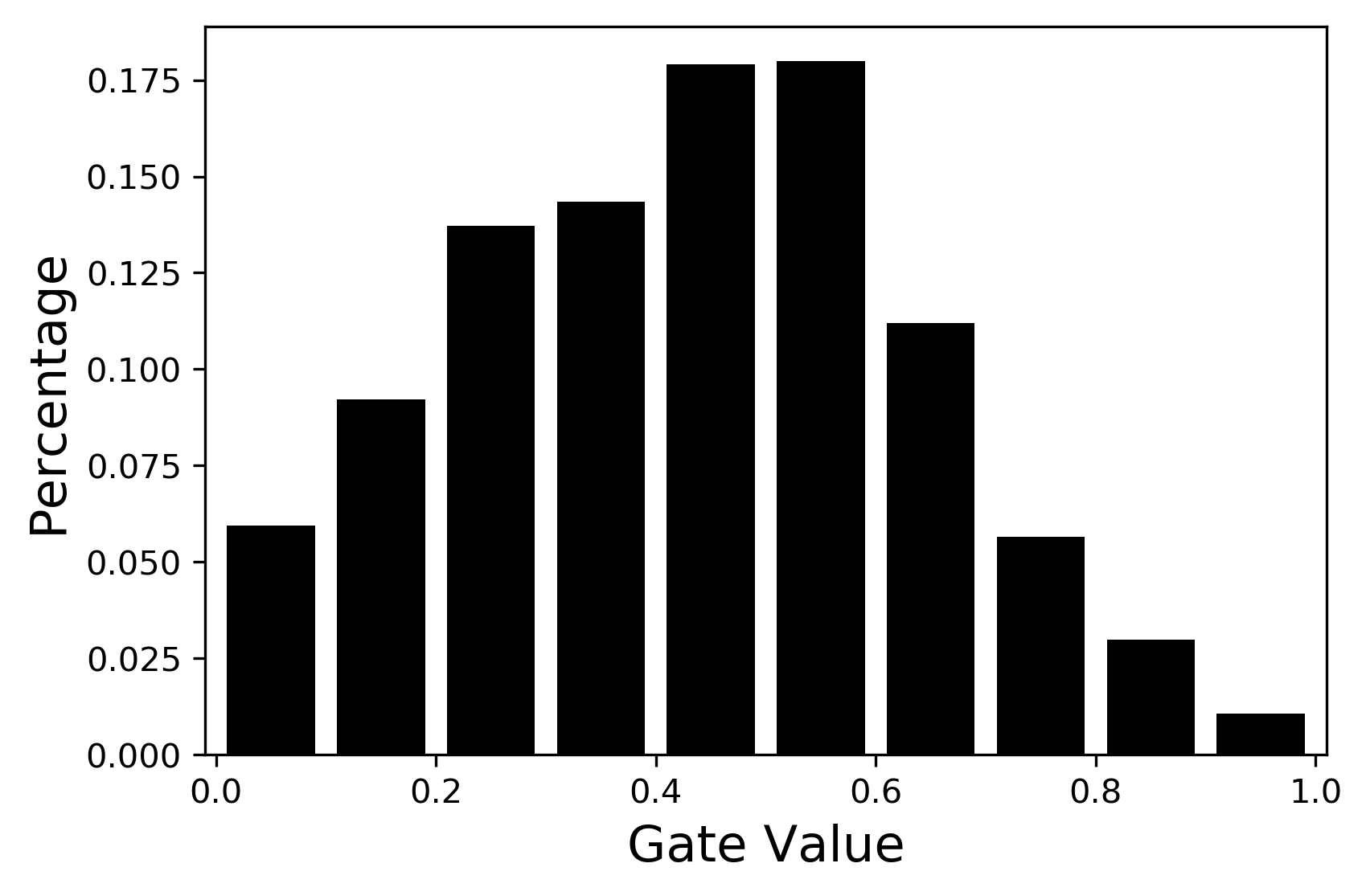}
		\caption{CIFG-LSTM} 
	\end{subfigure}\hfil
	\begin{subfigure}{.48\columnwidth}
		\centering
		\includegraphics[width=\linewidth]{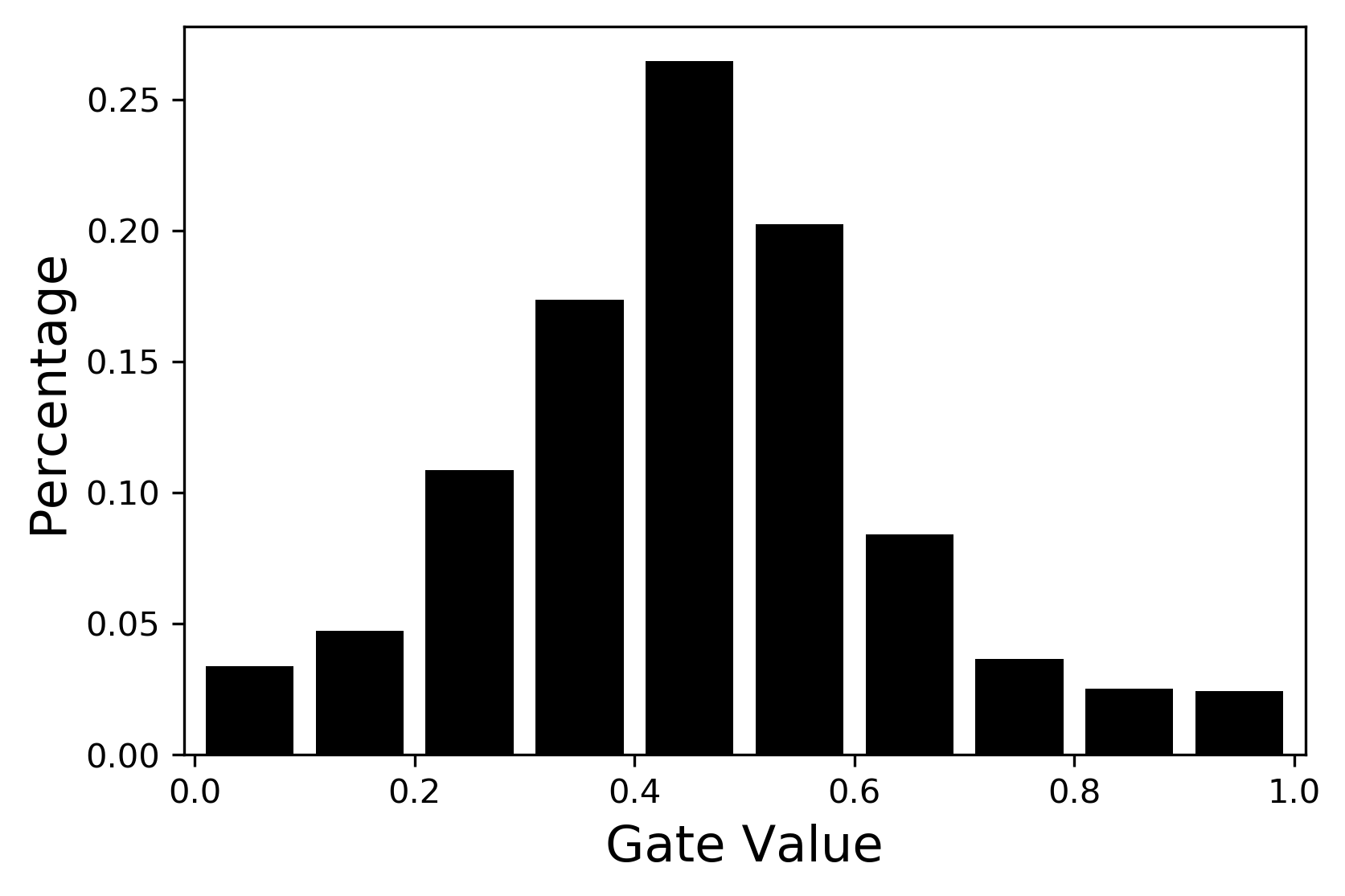}
		\caption{$G^{2}$-LSTM} 
	\end{subfigure}
	\begin{subfigure}{.48\columnwidth}
		\centering
		\includegraphics[width=\linewidth]{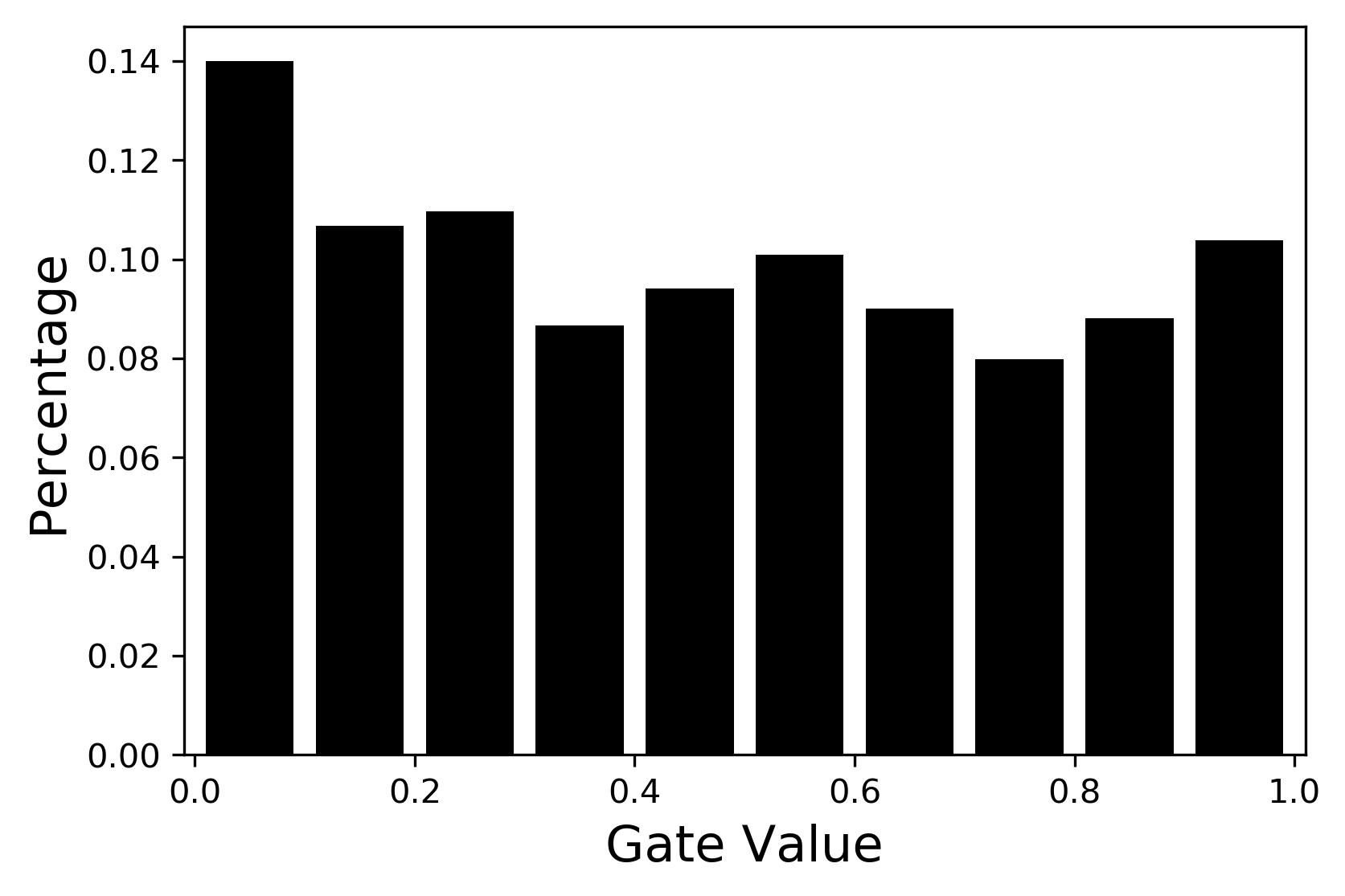}
		\caption{bBeta-LSTM(5G+p)}
	\end{subfigure}
	\caption{Histogram of input gate value on CR dataset. Our proposed model bBeta-LSTM(5G+p) shows the more flexible gate value than that of other models. CR dataset is used for the sentiment classification, and only a few words are important instead of whole words. As a result, the input gate in all models has a relatively higher portion of 0 than the portion of value 1. The bBeta-LSTM(5G+p) is more likely to have such a tendency, and it leads to better performance of bBeta-LSTM(5G+p) on CR dataset. 
%		(We provide the forget gate value in Appendix C.)
	}
	\label{fig:cr_gate_histogram}
\end{figure*}
\section{Experiments}
We compare our models and baselines, LSTM, CIFG-LSTM, $G^{2}$-LSTM, simple recurrent unit (SRU) \cite{sru}, R-transformer \cite{wang2019rtransf}, Batch normalized LSTM (BN-LSTM) \cite{cooijmans2016recurrent}, and h-detach \cite{hdetach2019}.
%\footnote{For the fair comparison, we use the $G^{2}$LSTM author's implementation code for $G^{2}$LSTM experiment https://github.com/zhuohan123/g2-lstm} on four benchmark tasks: text classifications, polyphonic music modeling, image classifications, and image captioning. 
First, we evaluate the performance of the bBeta-LSTM variants to measure the improvements from our structured gate modeling with the text classifications quantitatively and qualitatively on benchmark datasets. 
Second, we compare the models on polyphonic music modeling to check the performance of multi-label prediction tasks.
Third, we evaluate our models on a pixel-by-pixel MNIST dataset to confirm that our model can alleviate the gradient vanishing problems, empirically.
Finally, we perform the image caption generation task to check the performance on the multi-modal dataset.
\subsection{Text Classification}
%Text classification is one of the most frequent benchmark tasks for LSTM. To perform a sentence-level classification task, we need to model the overall semantics of a sentence by focusing on keywords. 
%For text classification, effective modeling on input and forget gates are necessary to attend the important context and to preserve the selective information, respectively.
We compare our models on six benchmark datasets, customer reviews (CR), sentence subjectivity (SUBJ), movie reviews (MR), question type (TREC), opinion polarity (MPQA), and Stanford Sentiment Treebank (SST). For LSTM models, we use a two-layer structure with 128 hidden dimensions for each layer, following \cite{sru}. We set the hidden dimensions of models to have the same number of parameters across the compared models 
%\footnote{\label{datastat} Appendix B provides the detailed statistical description for the datasets and the experimental settings.}.
%
Table \ref{table:sentence} shows the test accuracies for the model and dataset combinations. 
It should be noted that bBeta-LSTM(5G+p) performs better than other models on all datasets. In particular, bBeta-LSTM(5G) and bBeta-LSTM(5G+p), which provides an inductive bias of either positive or negative correlation, shows a significant improvement for CR dataset, the sparsest dataset in the benchmarks.
The performance difference between bBeta-LSTM(5G) and bBeta-LSTM(5G+p) shows the importance of the prior modeling to regularize the input and the forget gates.
\begin{figure*}[t!] %%%% Should be modified for better visualization
	\centering
	\includegraphics[width=\linewidth]{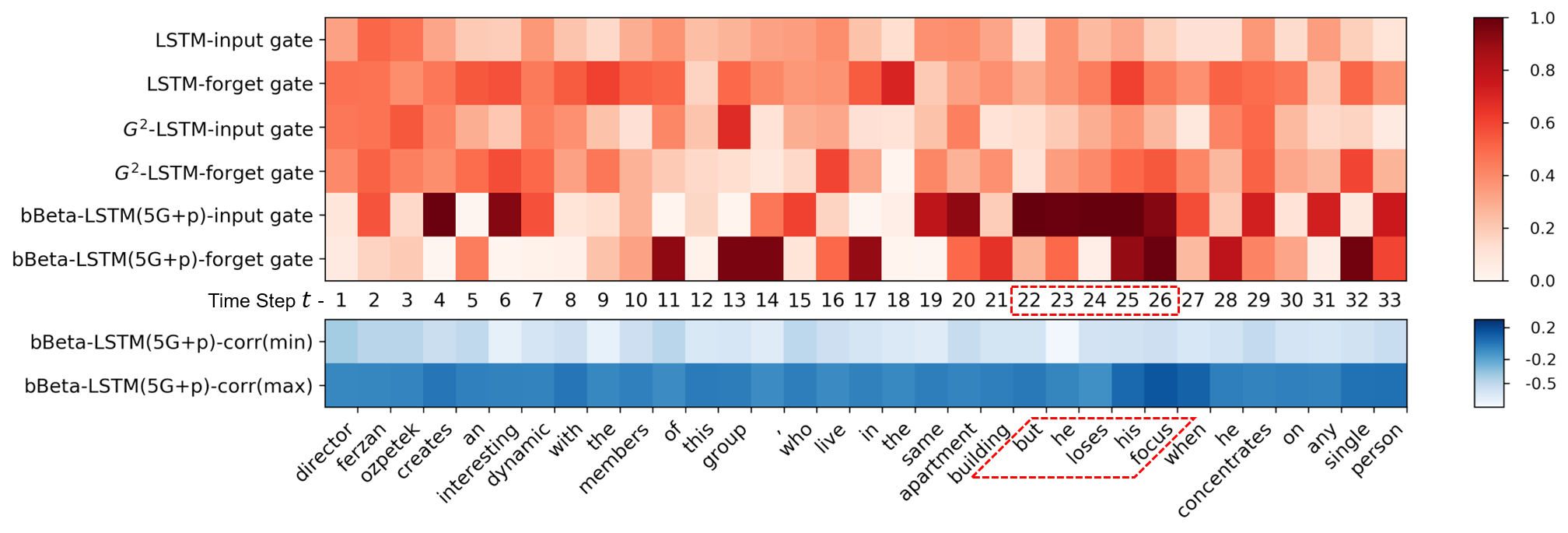}
	\caption{Visualization of input and forget gates for each model and the correlation for bBeta-LSTM(5G+p). The sentence with the negative label is designed for the sentiment classification task, and the "but he losses his focus"$t=22 \sim 26$ is an important part. At time step 22 ("but"), the change of context occurs, and bBeta-LSTM(5G+p) has a large input gate and relatively small forget gate to handle the context change. At time step 25 ("his"), both input gate and forget gate have high values to propagate the information "losses his" efficiently. This is the result of a relatively large correlation value at time step 25, and this correlation helps to propagate the information through the model.
%		(Appendix C provides the additional results)
	}
	\label{fig:case_study}
\end{figure*}
To check the compatibility with LSTM cell variants, we compare the performance between SRU and SRU+bBeta-LSTM(5G+p). For SRU+bBeta-LSTM(5G+p), we replace the gate structure in the SRU cell with our model, and it performs better than the original SRU for all datasets. SRU cell, which has two gate structures, is closer to GRU than LSTM, and this result demonstrates that our proposed gate structure can be compatible with other LSTM/GRU cell variants.
\begin{table*}[h!]
	\small
	\begin{minipage}{.61\linewidth}
		\centering
		\begin{tabular}{lcccc}
			\toprule
			Models & JSB & Muse & Nottingham & Piano \\
			\midrule
			LSTM & 8.68$\pm$0.10 & 7.17$\pm$0.06 &3.32$\pm$0.11 & 9.23$\pm$1.13 \\
			CIFG-LSTM & 8.69$\pm$0.06 & 7.18$\pm$0.01 &3.28$\pm$0.10 & 8.99$\pm$1.57 \\
			$G^{2}$-LSTM & 8.70$\pm$0.04 & 7.14$\pm$0.01 &3.23$\pm$0.04 & 9.00$\pm$0.84 \\
			\midrule
			Beta-LSTM   & 8.60$\pm$0.07 & 7.13$\pm$0.03& 3.30$\pm$0.06 & 8.24$\pm$0.26\\
			bBeta-LSTM(5G) & 8.63$\pm$0.12 & 7.11$\pm$0.04 & 3.30$\pm$0.04 & 8.43$\pm$0.64\\
			bBeta-LSTM(5G+p) & \textbf{8.30}$\pm$0.01 & \textbf{7.02}$\pm$0.02 & \textbf{3.14}$\pm$0.02 & \textbf{7.65}$\pm$0.08 \\
			\midrule
			R-Transformer   & 8.26$\pm$0.03 & 7.00$\pm$0.03& 2.24$\pm$0.01 & 7.44$\pm$0.03\\
			\hspace{1pt} + bBeta-LSTM(5G+p) & \textbf{8.24}$\pm$0.01 & \textbf{6.19}$\pm$0.02 & \textbf{2.13}$\pm$0.08 & \textbf{7.32}$\pm$0.03\\
			\bottomrule
		\end{tabular} 
		\caption{Negative log-likelihood on polyphonic music}
		\label{table:music}
	\end{minipage}%
	\begin{minipage}{.35\linewidth}
		\centering
		\begin{tabular}{lcc}
			\toprule
			Models & sMNIST & pMNIST \\
			\midrule
			LSTM    & 5.08$\pm$0.01 & 10.76$\pm$1.34 \\
			CIFG-LSTM & 1.23$\pm$0.13 & 8.42$\pm$0.58\\
			$G^{2}$-LSTM & 3.53$\pm$1.32 & 9.47$\pm$0.03\\
			\midrule
			Beta-LSTM   & 3.14$\pm$0.88 & 8.65$\pm$0.49 \\
			bBeta-LSTM(5G) & 1.75$\pm$0.50 & 8.37$\pm$0.46 \\
			bBeta-LSTM(5G+p) & \textbf{1.22}$\pm$0.25 & \textbf{7.66}$\pm$0.16 \\
			\midrule
			BN-LSTM & 1.05$\pm$0.06 & 4.26$\pm$0.50 \\
			\hspace{1pt} + bBeta-LSTM(5G+p) & \textbf{0.76}$\pm$0.05 & \textbf{3.90}$\pm$0.25 \\
			\bottomrule
		\end{tabular}
		\centering
		\caption{Test error rates on MNIST}
		\label{table:mnist}
	\end{minipage}
\end{table*}
We further examine the behavior of the bBeta-LSTM(5G+p) gate values from two perspectives of the input gate value ranges, and the input/forget correlations. 
Figure \ref{fig:prior_learning_corr_cr_mr} (Right) shows the correlation of bBeta-LSTM(5G+p), and the correlation between gates can exhibit both negative and positive values in bBeta-LSTM(5G+p).
Figure \ref{fig:cr_gate_histogram} visualizes the input gate and the forget gate values, and we observed that the input and the forget gate outputs fully utilize the range of [0,1] in bBeta-LSTM(5G+p). 

To further understand the model structure and its assumptions, we performed qualitative analysis on a sentence, which has a negative sentiment in the MR dataset.
Figure \ref{fig:case_study} shows the heatmap of the input gate and the forget gate for each model; and the correlation from our proposed model, bBeta-LSTM(5G+p). bBeta-LSTM(5G+p) model has a large input gate value on "but he loses his focus" ($t=22\sim26$) and a large forget gate value on 25 and 26 timestep to propagate the "losses" information well. Because of the structured gate modeling, bBeta-LSTM(5G+p) compose the meaning of "but he loses his focus" well. This effect originates from the structured gate modeling, which handles the correlation while other models do not model.
%, such as LSTM, $G^{2}$-LSTM, and Beta-LSTM, do not model.
There is a relatively large correlation in the "his focus" ($t=25,26$), and as a result, both input and forget gates have large values to propagate the important information efficiently. The sentiment label for the sentence is negative, and only bBeta-LSTM(5G+p) classifies it correctly.
%"director ferzan ozpetek creates an interesting dynamic with the members of this group, who live in the same apartment building, but he loses his focus when he concentrates on any single person."
\subsection{Polyphonic Music}
%Unlike the text classification whose purpose is predicting a single label for entire timesteps, polyphonic music modeling predicts a binary vector at every timestep. 
%Therefore, it is important to model the input and the forget gate appropriately for every timestep. 
We use four polyphonic music modeling benchmark datasets: JSB Chorales, Muse, Nottingham, and Piano. Table \ref{table:music} shows the test negative log-likelihood (NLL) on four music datasets.
%\cref{datastat}. 
%For the comparison, we use two-layered LSTM with 200 hidden dimensions for each layer with Adam optimizer. For a fair comparison, all models are adjusted to have the same number of parameters. 
Our proposed model, bBeta-LSTM(5G+p), performs better than all other models. 
%The number of training samples in the polyphonic music dataset is relatively small, and it shows that the prior modeling to the gate improves the robustness with a small dataset.
To compare with the pre-existing state-of-the-art model, we included the performance with R-Transformer \cite{wang2019rtransf} as well as R-Transformer with our gating mechanism. We replace the recurrent structure of R-transformer with our models, and our model shows better performance on all datasets. The results show high compatibility between our models and the Transformer model.

\subsection{Pixel by Pixel MNIST}
\begin{figure}[t!]
	\centering
	\begin{subfigure}{.48\columnwidth}
		\centering
		\includegraphics[width=\linewidth]{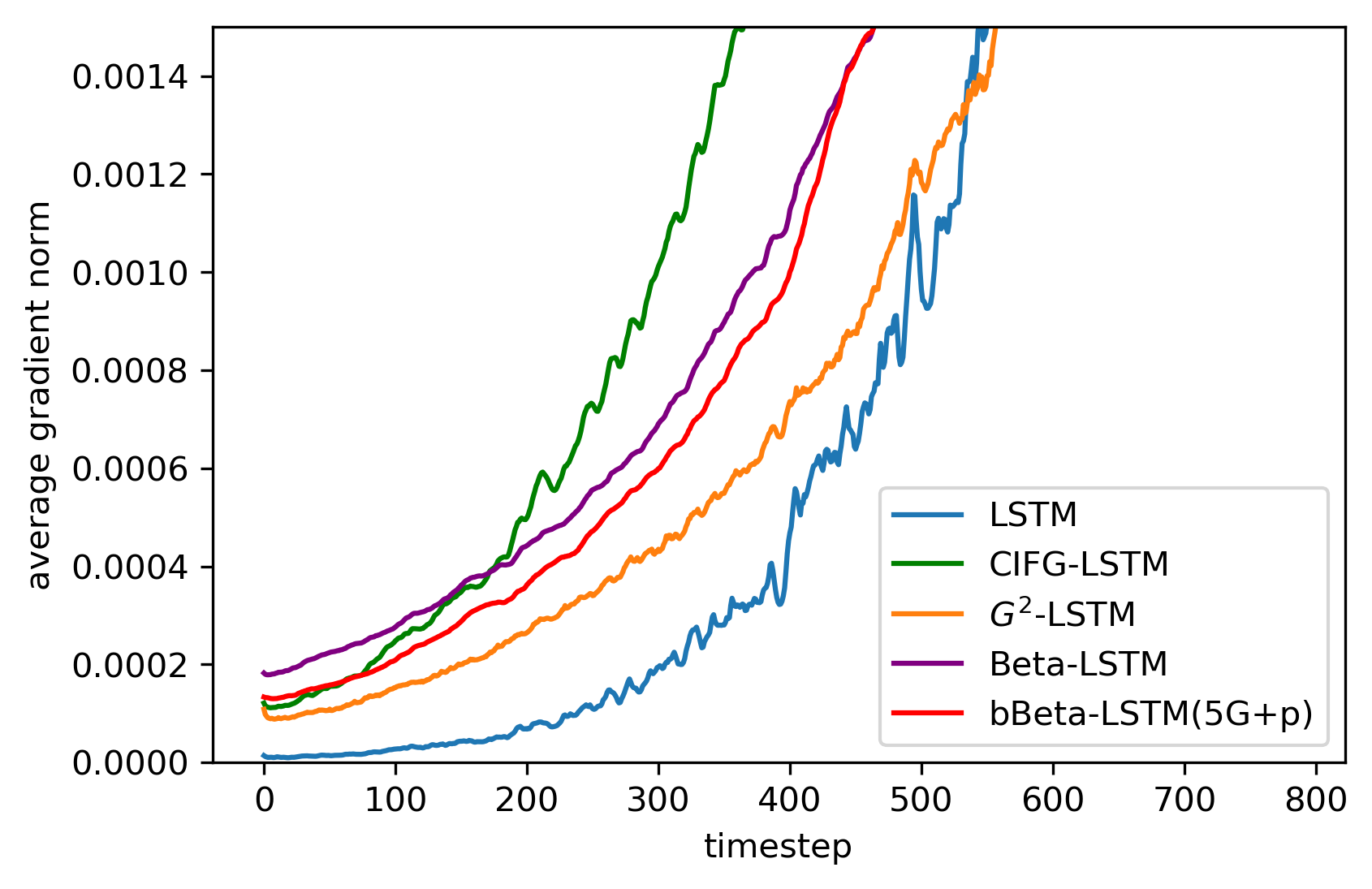}
		%		\caption{Average gradient norm on pMNIST dataset}
	\end{subfigure}
	\begin{subfigure}{.48\columnwidth}
		\centering
		\includegraphics[width=\linewidth]{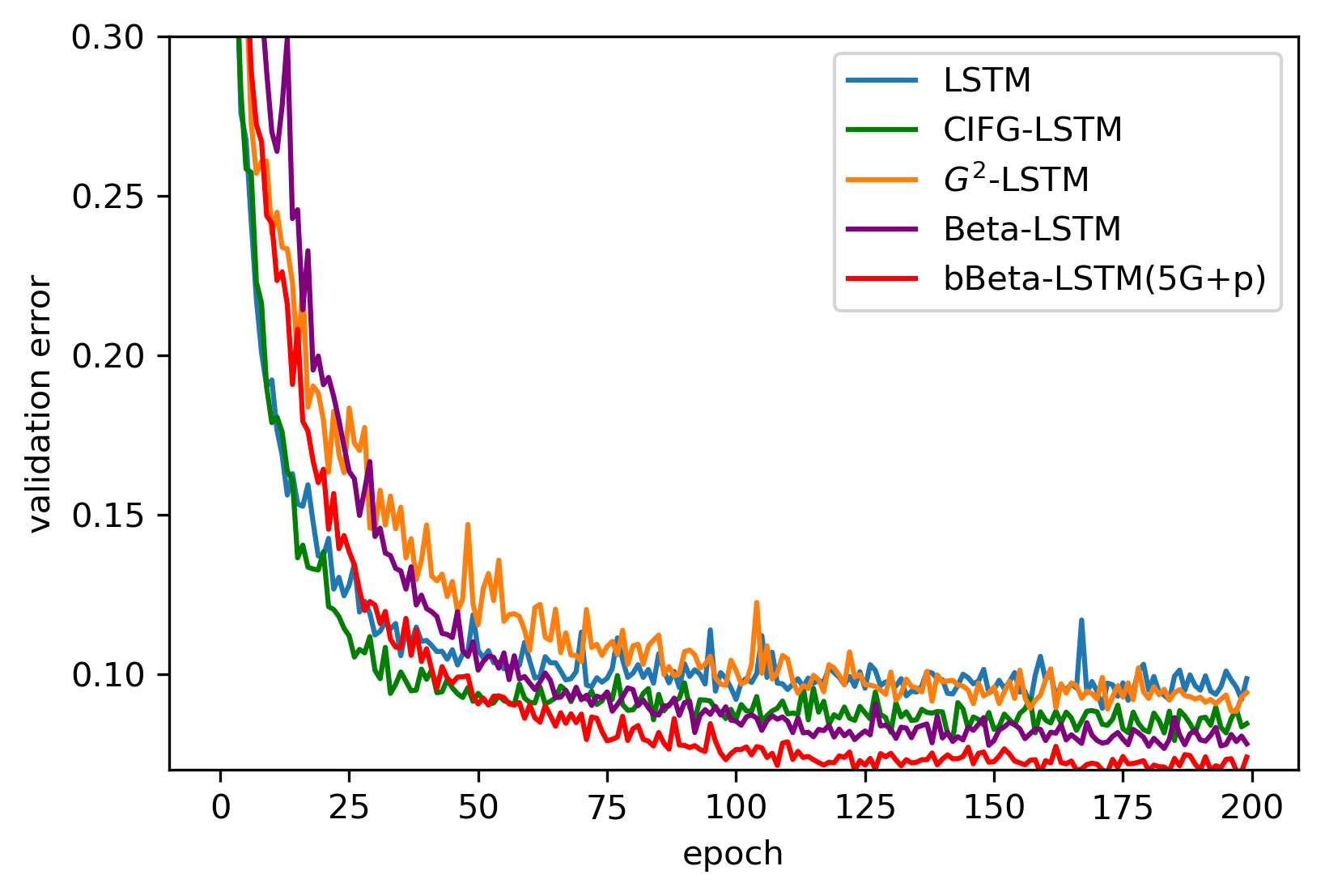}
		%		\caption{Validation curves pMNIST dataset}
	\end{subfigure}
	\caption{Average gradient norm, $\|{\frac{\partial L_{ELBO}}{\partial c_{t}}}\|$ for loss $L_{ELBO}$ over each time step. Beta-LSTM and bBeta-LSTM(5G+p) considers long-term dependency relatively well because they have larger gradients for initial timesteps, (left). bBeta-LSTM(5G+p), which incorporates the prior distribution, shows a relatively stable validation error curve, and shows the lowest validation error (right).}
	% between stocahstic models, $G^{2}$-LSTM, Beta-LSTM, and bBeta-LSTM(5G+p)}
	\label{fig:pMNIST_qualitative}
\end{figure}
\begin{table*}[h!]
	\centering
	\begin{tabular}{lcccccccc}
		\toprule
		\textbf{Models} & \textbf{B-1} & \textbf{B-2} & \textbf{B-3} & \textbf{B-4} & \textbf{METEOR} & \textbf{CIDEr} & \textbf{ROUGE-L} & \textbf{SPICE}\\
		\midrule
		DeepVS~\cite{karpathy2015deep} &62.5 & 45.0 & 32.1 & 23.0 & 19.5 & 66.0& {\textemdash}& {\textemdash} \\
		ATT-FCN~\cite{you2016image} & 70.9 & 53.7 & 40.2 & 30.4 & 24.3 & {\textemdash}& {\textemdash}& {\textemdash}\\
		Show \& Tell~\cite{vinyals2015show} & {\textemdash}  & {\textemdash}    & {\textemdash} & 27.7 & 23.7 & 85.5 & {\textemdash}& {\textemdash}\\  
		Soft Attention~\cite{xu2015show}  & 70.7 & 49.2 & 34.4 & 24.3 & 23.9 & {\textemdash}& {\textemdash}& {\textemdash}\\
		Hard Attention~\cite{xu2015show}  & 71.8 & 50.4 & 35.7 & 25.0 & 23.0 & {\textemdash}& {\textemdash}& {\textemdash}\\
		MSM~\cite{yao2017boosting} & 73.0 & 56.5 & 42.9 & 32.5 & 25.1 & 98.6 & {\textemdash}& {\textemdash}\\
		\midrule
		\multicolumn{7}{l}{\emph{Show\&Tell with Resnet152 (Our implementaion)}}& {\textemdash}& {\textemdash}\\
		\hspace{10pt}LSTM & 72.0 & 54.6 & 39.8 & 28.8 & 24.8 &  94.7& 52.5& 17.9\\
		\hspace{10pt}CIFG-LSTM & 71.2 & 53.9 & 39.3 & 28.5 & 24.4 & 93.0& 51.9& 17.7\\
		\hspace{10pt}$G^{2}$-LSTM & 71.7 & 54.3 & 39.7 & 28.8 & 24.6 & 93.0& 52.3& 17.5\\
		\hspace{10pt}bBeta-LSTM(5G+p) &  72.2 &  55.0 &  40.1 &  29.0 & 24.7 & 94.2& 52.6& 18.0\\
		\hspace{10pt}h-detach(0.4)~\cite{hdetach2019} & 72.2 & 55.0 & 40.9 & 30.3 & \bfseries 25.2 & 97.1 & 53.0& \bfseries 18.2\\
		\hspace{10pt}\hspace{1pt} + bBeta-LSTM(5G+p)  & \bfseries 72.3 & \bfseries 55.5 & \bfseries 41.5 & \bfseries 30.8 & \bfseries 25.2 & \bfseries 97.3 & \bfseries 53.2&  18.1\\
		\midrule
		\multicolumn{9}{l}{\emph{Show Attend Tell with Resnet152 (Our implementaion)}} \\
		\hspace{10pt}h-detach(0.4)~\cite{hdetach2019} & 73.3 & 56.7 & 42.6 & 31.8 & 25.8 & 101.2 & 54.0& 19.3\\
		\hspace{10pt}\hspace{1pt} + bBeta-LSTM(5G+p)  & \bfseries 74.1 & \bfseries 57.3 & \bfseries 43.1 & \bfseries 32.1 & \bfseries 26.1 & \bfseries 103.2 & \bfseries 54.3& \bfseries 19.4\\
		\bottomrule
	\end{tabular}
	\centering
	\caption{Test performance on MS-COCO dataset for BLEU, METEOR, CIDEr, ROUGE-L and SPICE evaluation metric.}
%	 We re-implement the \emph{Show \& Tell} and \emph{Show Attend Tell} based on resnet152, and compare the performance between models.
	\label{table:coco_results}
\end{table*}
The pixel-by-pixel MNIST task is predicting a category for a given 784 pixels. 
%There are two tasks for sequential MNIST: sMNIST, and pMNIST. 
sMNIST task handles each pixel with a sequential order, and pMNIST task models each pixel in a randomly permutated order. 
%784 timesteps are longer than the average text sentence length, so it becomes a challenging task to overcome the long term dependency. We divide the MNIST dataset into three sets: 50,000, 10,000, 10,000 for the train, the validation, and the test dataset, respectively. 
For the LSTM baseline, we use a single-layer model with 128 hidden dimensions with Adam optimizer.
%\cref{datastat}. 
Table \ref{table:mnist} shows the test error rates for sMNIST and pMNIST, and bBeta-LSTM(5G+p) shows the best performance. Besides, we compared the performance with the BN-LSTM, which performs well on sMNIST and pMNIST dataset.
When we replace the recurrent part in BN-LSTM with our model, we improve the test error rates about 27.6\% (from 1.05\% to 0.76\% error rate) in the pMNIST dataset. Batch normalization and its variants are important in various classification tasks, and the results show that our model is well compatible with the batch normalization methodology.
Left in Figure \ref{fig:pMNIST_qualitative} shows the gradients flow for each time step and the validation error curve for each epoch on the pMNIST dataset. For the gradient flow, we calculate the Frobenius norm of the gradient $\frac{\partial L_{ELBO}}{\partial c_{t}}$, and we average the norm over the image instance. We found that our proposed models, Beta-LSTM, and bBeta-LSTM(5G+p), propagate the information to the early timestep, efficiently.  
Right in Figure \ref{fig:pMNIST_qualitative} shows the validation error curve, and our proposed model bBeta-LSTM(5G+p), which incorporates the prior, shows the relatively stable learning curve.
% among the stochastic models.
%\begin{table*}[!htb]
\subsection{Image Captioning} % https://aclweb.org/anthology/E17-1019
%To verify the compatibility of our proposed model with other models, 
We evaluate our model on the image captioning task with Microsoft COCO dataset (MS-COCO) \cite{MSCOCO2014}. For the experiment, we split the dataset into 80,000, 5,000, 5,000 for the train, the validation, and the test dataset, respectively \cite{karpathy2015deep}. We use 512 hidden dimensions for the conditional caption generation, and we also used $Resnet152$ to retrieve image feature vectors.
%\cref{datastat}.
Table \ref{table:coco_results} shows the test performance for MS-COCO dataset based on $Show\&Tell$ \cite{vinyals2015show} and \emph{Show Attend Tell} \cite{xu2015show} encoder-decoder structure.
%bBeta-LSTM(5G+p) shows the best performance in BLEU, ROUGE-L and SPICE metric. 
Besides, to verify compatibility with other models, we re-implemented h-detach \cite{hdetach2019} and incorporate our models, bBeta-LSTM(5G+p).
When we replace the LSTM of h-detach with our models, we identified the improvement in the performance of h-detach.

\section{Conclusion}
We propose a new structured gate modeling which can improve the LSTM structure through the probabilistic modeling on gates. 
The gate structure in LSTM is a crucial component, and the gate value is the main controller for the information flow.
While the current sigmoid gate would satisfy the boundedness, we improve the sigmoid function with the Beta distribution to add flexibility. Moreover, bBeta-LSTM enables the detailed modeling of the covariance structure between gates, and bBeta-LSTM with prior guides the learning of the covariance structure. 
%The evaluations with benchmarks show the value of the probabilistic modeling, empirically. 
Also, our propositions state the improved characteristics of our probabilistic gate compared to the sigmoid function. 
From the application perspective, imposing the correlation between the input gate and the forget gate is necessary to handle the semantic information efficiently. 
This work envisions how to incorporate the neural network models with probabilistic components to improve its flexibility and stability. We demonstrated the necessity and effectiveness of flexible and prior modeling of gate structure on extensive experiments.

\subsubsection{Acknowledgments.}
This work was conducted at High-Speed Vehicle Research Center of KAIST with the support of the Defense Acquisition Program Administration and the Agency for Defense Development under Contract UD170018CD.

\clearpage
\bibliography{BetaLSTM}

\begin{thebibliography}{}

\bibitem[\protect\citeauthoryear{Arnold and Ng}{2011}]{ARNOLD20111194}
Arnold, B.~C., and Ng, H. K.~T.
\newblock 2011.
\newblock Flexible bivariate beta distributions.
\newblock {\em Journal of Multivariate Analysis} 102(8):1194--1202.

\bibitem[\protect\citeauthoryear{Battaglia \bgroup et al\mbox.\egroup
  }{2018}]{battaglia2018relational}
Battaglia, P.~W.; Hamrick, J.~B.; Bapst, V.; Sanchez-Gonzalez, A.; Zambaldi,
  V.; Malinowski, M.; Tacchetti, A.; Raposo, D.; Santoro, A.; Faulkner, R.;
  et~al.
\newblock 2018.
\newblock Relational inductive biases, deep learning, and graph networks.
\newblock {\em arXiv preprint arXiv:1806.01261}.

\bibitem[\protect\citeauthoryear{Blei, Ng, and Jordan}{2003}]{blei2003latent}
Blei, D.~M.; Ng, A.~Y.; and Jordan, M.~I.
\newblock 2003.
\newblock Latent dirichlet allocation.
\newblock {\em Journal of Machine Learning Research} 3:993--1022.

\bibitem[\protect\citeauthoryear{Chung \bgroup et al\mbox.\egroup
  }{2015}]{chung2015recurrent}
Chung, J.; Kastner, K.; Dinh, L.; Goel, K.; Courville, A.~C.; and Bengio, Y.
\newblock 2015.
\newblock A recurrent latent variable model for sequential data.
\newblock In {\em Advances in neural information processing systems},
  2980--2988.

\bibitem[\protect\citeauthoryear{Cooijmans \bgroup et al\mbox.\egroup
  }{2017}]{cooijmans2016recurrent}
Cooijmans, T.; Ballas, N.; Laurent, C.; G{\"u}l{\c{c}}ehre, {\c{C}}.; and
  Courville, A.
\newblock 2017.
\newblock Recurrent batch normalization.
\newblock {\em 5th International Conference on Learning Representations, {ICLR}
  2017,}.

\bibitem[\protect\citeauthoryear{Dieng \bgroup et al\mbox.\egroup
  }{2018}]{dieng2018noisin}
Dieng, A.~B.; Ranganath, R.; Altosaar, J.; and Blei, D.~M.
\newblock 2018.
\newblock Noisin: Unbiased regularization for recurrent neural networks.
\newblock In {\em Proceedings of the 35th International Conference on Machine
  Learning, {ICML} 2018, Stockholmsm{\"{a}}ssan, Stockholm, Sweden, July 10-15,
  2018},  1251--1260.

\bibitem[\protect\citeauthoryear{Gal and
  Ghahramani}{2016}]{gal2016theoretically}
Gal, Y., and Ghahramani, Z.
\newblock 2016.
\newblock A theoretically grounded application of dropout in recurrent neural
  networks.
\newblock In {\em Advances in neural information processing systems},
  1019--1027.

\bibitem[\protect\citeauthoryear{Greff \bgroup et al\mbox.\egroup
  }{2017}]{greff2017lstm}
Greff, K.; Srivastava, R.~K.; Koutn{\'\i}k, J.; Steunebrink, B.~R.; and
  Schmidhuber, J.
\newblock 2017.
\newblock Lstm: A search space odyssey.
\newblock {\em IEEE transactions on neural networks and learning systems}
  28(10):2222--2232.

\bibitem[\protect\citeauthoryear{Harabagiu}{2004}]{harabagiu2004incremental}
Harabagiu, S.
\newblock 2004.
\newblock Incremental topic representations.
\newblock In {\em Proceedings of the 20th international conference on
  Computational Linguistics},  583.
\newblock Association for Computational Linguistics.

\bibitem[\protect\citeauthoryear{Hochreiter and
  Schmidhuber}{1997}]{hochreiter1997long}
Hochreiter, S., and Schmidhuber, J.
\newblock 1997.
\newblock Long short-term memory.
\newblock {\em Neural computation} 9(8):1735--1780.

\bibitem[\protect\citeauthoryear{Jankowiak and
  Obermeyer}{2018}]{jankowiak2018pathwise}
Jankowiak, M., and Obermeyer, F.
\newblock 2018.
\newblock Pathwise derivatives beyond the reparameterization trick.
\newblock {\em arXiv preprint arXiv:1806.01851}.

\bibitem[\protect\citeauthoryear{Kampffmeyer \bgroup et al\mbox.\egroup
  }{2019}]{kampffmeyer2019connnet}
Kampffmeyer, M.; Dong, N.; Liang, X.; Zhang, Y.; and Xing, E.~P.
\newblock 2019.
\newblock Connnet: A long-range relation-aware pixel-connectivity network for
  salient segmentation.
\newblock {\em IEEE Transactions on Image Processing} 28(5):2518--2529.

\bibitem[\protect\citeauthoryear{Kanuparthi \bgroup et al\mbox.\egroup
  }{2019}]{hdetach2019}
Kanuparthi, B.; Arpit, D.; Kerg, G.; Ke, N.~R.; Mitliagkas, I.; and Bengio, Y.
\newblock 2019.
\newblock h-detach: Modifying the {LSTM} gradient towards better optimization.
\newblock In {\em 7th International Conference on Learning Representations,
  {ICLR} 2019, New Orleans, LA, USA, May 6-9, 2019}.

\bibitem[\protect\citeauthoryear{Karpathy and Li}{2015}]{karpathy2015deep}
Karpathy, A., and Li, F.
\newblock 2015.
\newblock Deep visual-semantic alignments for generating image descriptions.
\newblock In {\em {IEEE} Conference on Computer Vision and Pattern Recognition,
  {CVPR} 2015, Boston, MA, USA, June 7-12, 2015},  3128--3137.

\bibitem[\protect\citeauthoryear{Kingma and Welling}{2014}]{VAE2014}
Kingma, D.~P., and Welling, M.
\newblock 2014.
\newblock Auto-encoding variational bayes.
\newblock In {\em 2nd International Conference on Learning Representations,
  {ICLR} 2014, Banff, AB, Canada, April 14-16, 2014, Conference Track
  Proceedings}.

\bibitem[\protect\citeauthoryear{Lei \bgroup et al\mbox.\egroup }{2018}]{sru}
Lei, T.; Zhang, Y.; Wang, S.~I.; Dai, H.; and Artzi, Y.
\newblock 2018.
\newblock Simple recurrent units for highly parallelizable recurrence.
\newblock In {\em Proceedings of the 2018 Conference on Empirical Methods in
  Natural Language Processing},  4470--4481.
\newblock Brussels, Belgium: Association for Computational Linguistics.

\bibitem[\protect\citeauthoryear{Li \bgroup et al\mbox.\egroup
  }{2018}]{li2018towards}
Li, Z.; He, D.; Tian, F.; Chen, W.; Qin, T.; Wang, L.; and Liu, T.-Y.
\newblock 2018.
\newblock Towards binary-valued gates for robust lstm training.
\newblock {\em arXiv preprint arXiv:1806.02988}.

\bibitem[\protect\citeauthoryear{Lin \bgroup et al\mbox.\egroup
  }{2014}]{MSCOCO2014}
Lin, T.; Maire, M.; Belongie, S.~J.; Hays, J.; Perona, P.; Ramanan, D.;
  Doll{\'{a}}r, P.; and Zitnick, C.~L.
\newblock 2014.
\newblock Microsoft {COCO:} common objects in context.
\newblock In {\em Computer Vision - {ECCV} 2014 - 13th European Conference,
  Zurich, Switzerland, September 6-12, 2014, Proceedings, Part {V}},  740--755.

\bibitem[\protect\citeauthoryear{Olkin and Liu}{2003}]{OLKIN2003407}
Olkin, I., and Liu, R.
\newblock 2003.
\newblock A bivariate beta distribution.
\newblock {\em Statistics \& Probability Letters} 62(4):407--412.

\bibitem[\protect\citeauthoryear{Park \bgroup et al\mbox.\egroup
  }{2019}]{DBLP:journals/corr/abs-1904-09816}
Park, S.; Song, K.; Ji, M.; Lee, W.; and Moon, I.
\newblock 2019.
\newblock Adversarial dropout for recurrent neural networks.
\newblock {\em CoRR} abs/1904.09816.

\bibitem[\protect\citeauthoryear{Serban \bgroup et al\mbox.\egroup
  }{2017}]{serban2017hierarchical}
Serban, I.~V.; Sordoni, A.; Lowe, R.; Charlin, L.; Pineau, J.; Courville, A.;
  and Bengio, Y.
\newblock 2017.
\newblock A hierarchical latent variable encoder-decoder model for generating
  dialogues.
\newblock In {\em Thirty-First AAAI Conference on Artificial Intelligence}.

\bibitem[\protect\citeauthoryear{Vinyals \bgroup et al\mbox.\egroup
  }{2015}]{vinyals2015show}
Vinyals, O.; Toshev, A.; Bengio, S.; and Erhan, D.
\newblock 2015.
\newblock Show and tell: {A} neural image caption generator.
\newblock In {\em {IEEE} Conference on Computer Vision and Pattern Recognition,
  {CVPR} 2015, Boston, MA, USA, June 7-12, 2015},  3156--3164.

\bibitem[\protect\citeauthoryear{Wang \bgroup et al\mbox.\egroup
  }{2019}]{wang2019rtransf}
Wang, Z.; Ma, Y.; Liu, Z.; and Tang, J.
\newblock 2019.
\newblock R-transformer: Recurrent neural network enhanced transformer.
\newblock {\em arXiv preprint arXiv:1907.05572}.

\bibitem[\protect\citeauthoryear{Wolter and Yao}{2018}]{NIPS2018_8253}
Wolter, M., and Yao, A.
\newblock 2018.
\newblock Complex gated recurrent neural networks.
\newblock In Bengio, S.; Wallach, H.; Larochelle, H.; Grauman, K.;
  Cesa-Bianchi, N.; and Garnett, R., eds., {\em Advances in Neural Information
  Processing Systems 31}. Curran Associates, Inc.
\newblock  10536--10546.

\bibitem[\protect\citeauthoryear{Xu \bgroup et al\mbox.\egroup
  }{2015}]{xu2015show}
Xu, K.; Ba, J.; Kiros, R.; Cho, K.; Courville, A.~C.; Salakhutdinov, R.; Zemel,
  R.~S.; and Bengio, Y.
\newblock 2015.
\newblock Show, attend and tell: Neural image caption generation with visual
  attention.
\newblock In {\em Proceedings of the 32nd International Conference on Machine
  Learning, {ICML} 2015, Lille, France, 6-11 July 2015},  2048--2057.

\bibitem[\protect\citeauthoryear{Yao \bgroup et al\mbox.\egroup
  }{2017}]{yao2017boosting}
Yao, T.; Pan, Y.; Li, Y.; Qiu, Z.; and Mei, T.
\newblock 2017.
\newblock Boosting image captioning with attributes.
\newblock In {\em {IEEE} International Conference on Computer Vision, {ICCV}
  2017, Venice, Italy, October 22-29, 2017},  4904--4912.

\bibitem[\protect\citeauthoryear{You \bgroup et al\mbox.\egroup
  }{2016}]{you2016image}
You, Q.; Jin, H.; Wang, Z.; Fang, C.; and Luo, J.
\newblock 2016.
\newblock Image captioning with semantic attention.
\newblock In {\em 2016 {IEEE} Conference on Computer Vision and Pattern
  Recognition, {CVPR} 2016, Las Vegas, NV, USA, June 27-30, 2016},  4651--4659.

\end{thebibliography}
\bibliographystyle{aaai}
\end{document}